# Generating Approximate Solutions to the Traveling Tournament Problem using a Linear Distance Relaxation

**Richard Hoshino** RICHARD.HOSHINO@GMAIL.COM
**Ken-ichi Kawarabayashi** K_KENITI@NII.AC.JP
*National Institute of Informatics and*
*JST ERATO Kawarabayashi Project*
*2-1-2 Hitotsubashi, Chiyoda-ku, Tokyo 101-8430, Japan*

## Abstract

In some domestic professional sports leagues, the home stadiums are located in cities connected by a common train line running in one direction. For these instances, we can incorporate this geographical information to determine optimal or nearly-optimal solutions to the $n$-team Traveling Tournament Problem (TTP), an NP-hard sports scheduling problem whose solution is a double round-robin tournament schedule that minimizes the sum total of distances traveled by all $n$ teams.

We introduce the Linear Distance Traveling Tournament Problem (LD-TTP), and solve it for $n = 4$ and $n = 6$, generating the complete set of possible solutions through elementary combinatorial techniques. For larger $n$, we propose a novel "expander construction" that generates an approximate solution to the LD-TTP. For $n \equiv 4 \pmod 6$, we show that our expander construction produces a feasible double round-robin tournament schedule whose total distance is guaranteed to be no worse than $\frac{4}{3}$ times the optimal solution, regardless of where the $n$ teams are located. This $\frac{4}{3}$-approximation for the LD-TTP is stronger than the currently best-known ratio of $\frac{5}{3} + \epsilon$ for the general TTP.

We conclude the paper by applying this linear distance relaxation to general (non-linear) $n$-team TTP instances, where we develop fast approximate solutions by simply "assuming" the $n$ teams lie on a straight line and solving the modified problem. We show that this technique surprisingly generates the distance-optimal tournament on all benchmark sets on 6 teams, as well as close-to-optimal schedules for larger $n$, even when the teams are located around a circle or positioned in three-dimensional space.

## 1. Introduction

In this paper, we introduce a simple yet powerful technique to develop approximate solutions to the Traveling Tournament Problem (TTP), by "assuming" the $n$ teams are located on a straight line, thereby reducing the $\binom{n}{2}$ pairwise distance parameters to just $n-1$ variables, and then solving the relaxed problem.

The Traveling Tournament Problem (TTP) was inspired by the real-life problem of optimizing the regular-season schedule for Major League Baseball. The goal of the TTP is to determine the optimal double round-robin tournament schedule for an $n$-team sports league that minimizes the sum total of distances traveled by all $n$ teams. Since the problem was first proposed (Easton, Nemhauser, & Trick, 2001), the TTP has attracted a significant amount of research (Kendall, Knust, Ribeiro, & Urrutia, 2010), with numerous heuristics developed for solving hard TTP instances, such as local search techniques as well as integer and constraint programming.

In many ways, the TTP is a variant of the well-known Traveling Salesman Problem (TSP), asking for a distance-optimal schedule linking venues that are close to one another. The computational complexity of the TSP is NP-hard; recently, it was shown that solving the TTP is strongly NP-hard (Thielen & Westphal, 2010).

In the *Linear Distance* Traveling Tournament Problem (LD-TTP), we assume the $n$ teams are located on a straight line. This straight line relaxation is a natural heuristic when the $n$ teams are located in cities connected by a common train line running in one direction, modelling the actual context of domestic sports leagues in countries such as Chile, Sweden, Italy, and Japan. For example, Figure 1 illustrates the locations of the six home stadiums in Nippon Pro Baseball's Central League, all situated in close proximity to major stations on Japan's primary bullet-train line.

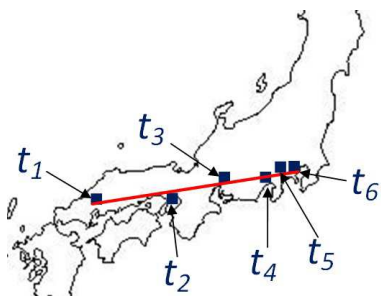

Figure 1: The six Central League teams in Japanese Pro Baseball.

In Section 2, we formally define the TTP. In Section 3, we solve the LD-TTP for $n = 4$ and list all 18 non-isomorphic tournament schedules achieving the optimal distance. In Section 4, we solve the LD-TTP for $n = 6$ and show that there are 295 non-isomorphic tournament schedules that can attain one of the seven possible values for the optimal distance. In Section 5, we provide an "expander construction" to produce a feasible double round-robin tournament schedule for any tournament on $n = 6m - 2$ teams, and prove that it is a $\frac{4}{3}$-approximation of the distance-optimal schedule, for any $m \geq 1$. In Section 6, we apply our theories to all known (non-linear) 6-team benchmark sets (Trick, 2012), and show that in all cases, the optimal solution appears in our list of 295. We also apply our expander construction to various benchmark sets on 10 and 16 teams, showing that this optimality gap is actually far lower than the theoretical maximum of 33.3%. In Section 7, we do an in-depth analysis of the optimality gap, and conclude the paper in Section 8 with some open problems and directions for future research.

## 2. The Traveling Tournament Problem

Let $\{t_1, t_2, \ldots, t_n\}$ be the $n$ teams in a sports league, where $n$ is even. Let $D$ be the $n \times n$ *distance matrix*, where entry $D_{i,j}$ is the distance between the home stadiums of teams $t_i$ and $t_j$. By definition, $D_{i,j} = D_{j,i}$ for all $1 \leq i, j \leq n$, and all diagonal entries $D_{i,i}$ are zero. We assume the distances form a *metric*, i.e., $D_{i,j} \leq D_{i,k} + D_{k,j}$ for all $i, j, k$.

The TTP requires a tournament lasting $2(n-1)$ days, where every team has exactly one game scheduled each day with no byes or days off (this explains why $n$ must be even.) The objective is to minimize the total distance traveled by the $n$ teams, subject to the following conditions:

(a) *each-venue*: Each pair of teams plays twice, once in each other's home venue.

(b) *at-most-three*: No team may have a home stand or road trip lasting more than three games.

(c) *no-repeat*: A team cannot play against the same opponent in two consecutive games.

When calculating the total distance, we assume that every team begins the tournament at home and returns home after playing its last away game. Furthermore, whenever a team has a road trip consisting of multiple away games, the team doesn't return to their home city but rather proceeds directly to their next away venue.

To illustrate with a specific example, Table 1 lists the distance-optimal schedule (Easton et al., 2001) for a bechmark set known as NL6 (six teams from Major League Baseball's National League). In this schedule, as with all subsequent schedules presented in this paper, home games are marked in bold.

| Team | 1 | 2 | 3 | 4 | 5 | 6 | 7 | 8 | 9 | 10 |
|---|---|---|---|---|---|---|---|---|---|---|
| Florida (FLO) | ATL | PHI | **NYK** | **PIT** | NYK | MON | PIT | **PHI** | **MON** | **ATL** |
| Atlanta (ATL) | **FLO** | **NYK** | **PIT** | PHI | MON | PIT | **PHI** | **MON** | NYK | FLO |
| Pittsburgh (PIT) | **NYK** | **MON** | ATL | FLO | **PHI** | **ATL** | **FLO** | NYK | PHI | MON |
| Philadelphia (PHI) | MON | **FLO** | MON | **ATL** | PIT | **NYK** | ATL | FLO | **PIT** | NYK |
| New York (NYK) | PIT | ATL | FLO | **MON** | **FLO** | PHI | MON | **PIT** | **ATL** | **PHI** |
| Montreal (MON) | **PHI** | PIT | **PHI** | NYK | **ATL** | **FLO** | **NYK** | ATL | FLO | **PIT** |

Table 1: An optimal TTP solution for NL6.

For example, the total distance traveled by Florida is $D_{\text{FLO,ATL}} + D_{\text{ATL,PHI}} + D_{\text{PHI,FLO}} + D_{\text{FLO,NYK}} + D_{\text{NYK,MON}} + D_{\text{MON,PIT}} + D_{\text{PIT,FLO}}$. Based on the NL6 distance matrix (Trick, 2012), the tournament schedule in Table 1 requires 23916 miles of total team travel, which can be shown to be the minimum distance possible.

## 3. The 4-Team LD-TTP

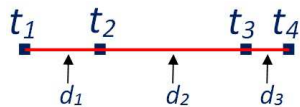

Figure 2: The general instance of the LD-TTP for $n = 4$.

In the Linear Distance TTP, we assume the $n$ home stadiums lie on a straight line, with $t_1$ at one end and $t_n$ at the other. Thus, $D_{i,j} = D_{i,k} + D_{k,j}$ for all triplets $(i, j, k)$ with

259

$1 \leq i < k < j \leq n$. Since the Triangle Inequality is replaced by the Triangle Equality, we no longer need to consider all $\binom{n}{2}$ entries in the distance matrix $D$; each tournament's total travel distance is a function of $n-1$ variables, namely the set $\{D_{i,i+1} : 1 \leq i \leq n-1\}$. For notational convenience, denote $d_i := D_{i,i+1}$ for all $1 \leq i \leq n-1$.

| Team | 1 | 2 | 3 | 4 | 5 | 6 |
|---|---|---|---|---|---|---|
| $t_1$ | $\boldsymbol{t_4}$ | $\boldsymbol{t_3}$ | $t_2$ | $t_4$ | $t_3$ | $\boldsymbol{t_2}$ |
| $t_2$ | $t_3$ | $t_4$ | $\boldsymbol{t_1}$ | $t_3$ | $t_4$ | $t_1$ |
| $t_3$ | $t_2$ | $t_1$ | $\boldsymbol{t_4}$ | $\boldsymbol{t_2}$ | $\boldsymbol{t_1}$ | $t_4$ |
| $t_4$ | $t_1$ | $t_2$ | $t_3$ | $\boldsymbol{t_1}$ | $\boldsymbol{t_2}$ | $\boldsymbol{t_3}$ |

Table 2: An optimal LD-TTP solution for $n = 4$.

Table 2 gives a feasible solution to the 4-team LD-TTP. We claim that this solution is optimal, for all possible 3-tuples $(d_1, d_2, d_3)$. To see why this is so, define $ILB_{t_i}$ to be the *independent lower bound* for team $t_i$, the minimum possible distance that can be traveled by $t_i$ in order to complete its games, independent of the other teams' schedules. Then a trivial lower bound for the total travel distance is $TLB \geq \sum_{i=1}^{n} ILB_{t_i}$.

Recall that when calculating $t_i$'s travel distance, we assume that $t_i$ begins the tournament at home and returns home after playing its last away game. Since $t_i$ must play a road game against each of the other three teams, $ILB_{t_i} = 2(d_1 + d_2 + d_3)$ for all $1 \leq i \leq 4$. This implies that $TLB \geq 8(d_1 + d_2 + d_3)$. Since Table 2 is a tournament schedule whose total distance is the trivial lower bound, this completes the proof.

We remark that Table 2 is not the unique solution - for example, we can generate another optimal schedule by simply reading Table 2 from right to left. Assuming the first match between $t_1$ and $t_2$ occurs in the home city of $t_2$, a straightforward computer search finds 18 different schedules with total distance $8(d_1 + d_2 + d_3)$, which are provided in Table 3 below. (For readability, we have replaced each occurrence of $t_i$ by the single index $i$.) Thus, by symmetry, there are 36 optimal schedules for the 4-team LD-TTP. For the interested reader, Appendix A provides the actual Maplesoft code that generated these optimal schedules.

```
234234  234243  234342  243234  243243  243432  342342  342432  432342
143143  143134  143431  134143  134134  134341  431431  431341  341431
412412  412421  412124  421412  421421  421214  124124  124214  214124
321321  321312  321213  312321  312312  312123  213213  213123  123213

432432  342342  342432  342342  342432  432342  432432  432342  432432
341341  431431  431341  431431  431341  341431  341341  341431  341341
214214  124124  124214  124124  124214  214124  214214  214124  214214
123123  213213  213123  213213  213123  123213  123123  123213  123123
```

Table 3: The eighteen non-isomorphic optimal LD-TTP solutions for $n = 4$.

This completes the analysis of the 4-team Linear Distance TTP.

## 4. The 6-Team LD-TTP

Unlike the previous section, the analysis for the 6-team LD-TTP requires more work.

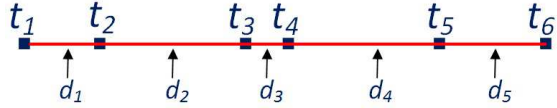

Figure 3: The general instance of the LD-TTP for $n = 6$.

Any 6-team instance of the LD-TTP can be represented by the 5-tuple $(d_1, d_2, d_3, d_4, d_5)$. We define $S = 14d_1 + 16d_2 + 20d_3 + 16d_4 + 14d_5$. We claim the following:

**Theorem 1.** *Let $\Gamma$ be a 6-team instance of the LD-TTP. The optimal solution to $\Gamma$ is a schedule with total distance $S + 2\min\{d_2 + d_4, d_1 + d_4, d_3 + d_4, 3d_4, d_2 + d_5, d_2 + d_3, 3d_2\}$.*

We will prove Theorem 1 through elementary combinatorial arguments, thus demonstrating the utility of this linear distance relaxation and presenting new techniques to attack the general TTP in ways that differ from integer/constraint programming. Our proof will follow from several lemmas, which we now prove one by one.

**Lemma 1.** *Any feasible schedule of $\Gamma$ must have total distance at least $S$.*

*Proof.* For each $1 \leq k \leq 5$, define $c_k$ to be the total number of times a team crosses the "bridge" of length $d_k$, connecting the home stadiums of teams $t_k$ and $t_{k+1}$. Let $Z$ be the total travel distance of this schedule. Since $\Gamma$ is linear, $Z = \sum_{k=1}^{5} c_k d_k$. Since each team crosses every bridge an even number of times, $c_k$ is always even.

Let $L_k$ be the home venues of $\{t_1, t_2, \ldots, t_k\}$ and $R_k$ be the home venues of $\{t_{k+1}, \ldots, t_6\}$. By the *each-venue* condition, every team in $L_k$ plays a road game against every team in $R_k$. By the *at-most-three* condition, every team in $L_k$ must make at least $2\lceil \frac{6-k}{3} \rceil$ trips across the bridge, with half the trips in each direction. Similarly, every team in $R_k$ must make at least $2\lceil \frac{k}{3} \rceil$ trips across the bridge, implying that $c_k \geq 2k\lceil \frac{6-k}{3} \rceil + 2(6-k)\lceil \frac{k}{3} \rceil$.

Thus, $c_1 \geq 14$, $c_2 \geq 16$, $c_4 \geq 16$, and $c_5 \geq 14$. We now show that $c_3 \geq 20$, which will complete the proof that $Z = \sum c_k d_k \geq 14d_1 + 16d_2 + 20d_3 + 16d_4 + 14d_5 = S$.

Since there are $n = 6$ teams, there are $2(n-1) = 10$ days of games. For each $1 \leq i \leq 9$, let $X_{i,i+1}$ be the total number of times the $d_3$-length bridge is crossed as the teams move from their games on the $i^{\text{th}}$ day to their games on the $(i+1)^{\text{th}}$ day. Let $X_{start,1}$ and $X_{10,end}$ respectively be the number of times the teams cross this bridge to play their first game, and return home after having played their last game. Then $c_3 = X_{start,1} + \sum_{i=1}^{9} X_{i,i+1} + X_{10,end}$.

For each $1 \leq i \leq 9$, let $f(i)$ denote the number of games played in $L_3$ on day $i$. Thus, on day $i$, exactly $2f(i)$ teams are to the left of this bridge and $6 - 2f(i)$ teams are to the right. So $f(i) \in \{0, 1, 2, 3\}$ for all $i$. Since $|L_3|$ and $|R_3|$ are odd, we have $X_{start,1} \geq 1$ and $X_{10,end} \geq 1$.

If $f(i) < f(i+1)$, then $X_{i,i+1} \geq 2$, as at least two teams who played in $R_3$ on day $i$ must cross over to play their next game in $L_3$. Similarly, if $f(i) > f(i+1)$, then $X_{i,i+1} \geq 2$.

If $f(i) = f(i+1) = 1$, then on day $i$, two teams $p$ and $q$ play in $L_3$ while the other four teams play in $R_3$. If $X_{i,i+1} = 0$ then no team crosses the bridge after day $i$, forcing $p$ and

261

$q$ to play against each other on day $i+1$, thus violating the *no-repeat* condition. Thus, at least one of $p$ or $q$ must cross the bridge, exchanging positions with at least one other team who must cross to play in $L_3$. Thus, $X_{i,i+1} \geq 2$. Similarly, if $f(i) = f(i+1) = 2$, then $X_{i,i+1} \geq 2$.

If $f(i) = f(i+1) = 0$, then all teams play in $R_3$ on days $i$ and $i+1$. Then $X_{start,1} = 3$ if $i = 1$ and $X_{10,end} = 3$ if $i = 9$. If $2 \leq i \leq 8$, then each of $\{t_1, t_2, t_3\}$ must play a home game on either day $i-1$ or day $i+2$, in order to satisfy the *at-most-three* condition. Thus, on one of these two days, at least two teams in $\{t_1, t_2, t_3\}$ play at home, implying at least four teams are in $L_3$. Therefore, we must have $X_{i-1,i} \geq 4$ or $X_{i+1,i+2} \geq 4$.

We derive the same results if $f(i) = f(i+1) = 3$. We have $X_{start,1} = 3$ if $i = 1$, $X_{10,end} = 3$ if $i = 9$, and either $X_{i-1,i} \geq 4$ or $X_{i+1,i+2} \geq 4$ if $2 \leq i \leq 8$.

So in our double round-robin schedule, if the sequence $\{f(1), \ldots, f(10)\}$ has no pair of consecutive 0s or consecutive 3s, then $c_3 = X_{start,1} + \sum_{i=1}^{9} X_{i,i+1} + X_{10,end} \geq 1 + 9 \cdot 2 + 1 = 20$. And if this is not the case, we still have $c_3 \geq 20$ from the results of the previous two paragraphs. We have therefore proven that $Z = \sum c_k d_k \geq S$. □

**Lemma 2.** *Consider a feasible schedule of $\Gamma$ with total distance $Z = \sum c_k d_k$. If $c_2 = 16$, then teams $t_1$ and $t_2$ must play against each other on Days 1 and 10.*

*Proof.* As we did in Lemma 1, for each $1 \leq i \leq 9$ define $X^*_{i,i+1}$ be the total number of times the $d_2$-length bridge is crossed as the teams move from their games on the $i^{\text{th}}$ day to their games on the $(i+1)^{\text{th}}$ day. Similarly define $X^*_{start,1}$ and $X^*_{10,end}$ so that $c_2 = X^*_{start,1} + \sum_{i=1}^{9} X^*_{i,i+1} + X^*_{10,end}$.

We now prove that $\sum_{i=1}^{9} X^*_{i,i+1} \geq 16$. To do this, for each $1 \leq i \leq 10$, let $g(i)$ denote the number of games played in $L_2$ (i.e., the home stadiums of $t_1$ and $t_2$) on day $i$. So on day $i$, exactly $2g(i)$ teams are to the left of the $d_2$-length bridge and $6 - 2g(i)$ teams are to the right. Clearly, $0 \leq g(i) \leq 2$ for all $1 \leq i \leq 10$.

If $|g(i+1) - g(i)| = 1$, then $X^*_{i,i+1} \geq 2$, as at least two teams who played on day $i$ on one side of the bridge must cross over to play their next game on the other side. If $|g(i+1) - g(i)| = 2$, then $X^*_{i,i+1} = 4$.

If $g(i) = g(i+1) = 1$, then on day $i$, two teams $p$ and $q$ play in $L_2$ while the other four teams play in $R_2$. If $X^*_{i,i+1} = 0$ then no team crosses the bridge after day $i$, forcing $p$ and $q$ to play against each other on day $i+1$, thus violating the *no-repeat* condition. Thus, at least one of $p$ or $q$ must cross the bridge, exchanging positions with at least one other team who must cross to play in $L_2$. Thus, $X^*_{i,i+1} \geq 2$. Similarly, if $g(i) = g(i+1) = 2$, then two teams $p$ and $q$ play in $R_2$ while the other four teams play in $L_2$, and we apply the same argument to show that $X^*_{i,i+1} \geq 2$. The remaining case to consider is $g(i) = g(i+1) = 0$, in which case $X^*_{i,i+1}$ could equal 0.

Suppose there are $a$ days with $g(i) = 0$, $b$ days with $g(i) = 1$, and $c$ days with $g(i) = 2$. Then $a + b + c = 10$. Since each of $t_1$ and $t_2$ play five home games, this implies $b + 2c = 10$. From this, we see that $a = c$ and that there are only six possible triplets for $(a, b, c)$, namely $(0, 10, 0)$, $(1, 8, 1)$, $(2, 6, 2)$, $(3, 4, 3)$, $(4, 2, 4)$, and $(5, 0, 5)$.

If $a = 0$ or $a = 1$, then there does not exist an index $i$ with $g(i) = g(i+1) = 0$, implying that $X^*_{i,i+1} \geq 2$ for all $1 \leq i \leq 9$. Hence, $\sum_{i=1}^{9} X^*_{i,i+1} \geq 9 \times 2 = 18$ in these cases. If $a = 2$, then there is at most one index $i$ with $g(i) = g(i+1) = 0$, implying that $\sum_{i=1}^{9} X^*_{i,i+1} \geq 16$.

Suppose $\sum_{i=1}^{9} X_{i,i+1}^* < 16$. Then we must have $3 \leq a \leq 5$, with two or more indices $i$ satisfying $g(i) = g(i+1) = 0$. For example, one such 10-tuple is $(g(1), g(2), \ldots, g(9), g(10)) = (1, 0, 0, 0, 1, 1, 1, 2, 2, 2)$, which can have $\sum_{i=1}^{9} X_{i,i+1}^* = 14$. A simple case analysis of each of $(a, b, c) \in \{(3, 4, 3), (4, 2, 4), (5, 0, 5)\}$ shows that all such "bad" 10-tuples violate the at-most-three condition; for example, in the 10-tuple above, either $t_1$ or $t_2$ must play four consecutive road games to start the tournament, which is a contradiction.

Therefore, we have proven that $\sum_{i=1}^{9} X_{i,i+1}^* \geq 16$ in all cases. This implies that if $c_2 = 16$, then $X_{start,1}^* = X_{10,end}^* = 0$. Hence, on Days 1 and 10, $t_1$ and $t_2$ stay in $L_2$ while the other four teams stay in $R_2$. Since $t_1$ and $t_2$ are the only teams in $L_2$, clearly this forces these two teams to play against each other, to begin and end the tournament. □

**Lemma 3.** *Let $S_1$ be the set of tournament schedules with distance $S + 2(d_2 + d_4)$, $S_2$ with distance $S + 2(d_1 + d_4)$, $S_3$ with distance $S + 2(d_3 + d_4)$, $S_4$ with distance $S + 6d_4$, $S_5$ with distance $S + 2(d_2 + d_5)$, $S_6$ with distance $S + 2(d_2 + d_3)$, and $S_7$ with distance $S + 6d_2$. Then each set in $\{S_1, S_2, \ldots, S_7\}$ is non-empty.*

*Proof.* For each of these seven sets, it suffices to find just one feasible schedule with the desired total distance. For each of $\{S_1, S_2, S_3, S_4\}$, at least one such set has appeared previously in the literature, as the solution to a 6-team benchmark set or in some other context. (As we will see in the following section, we can label the six teams of the NL6 benchmark set so that Table 1 is an element of $S_4$.)

|       | 1     | 2     | 3     | 4         | 5         | 6     | 7         | 8         | 9         | 10    | $d_1$ | $d_2$ | $d_3$ | $d_4$ | $d_5$ |
|-------|-------|-------|-------|-----------|-----------|-------|-----------|-----------|-----------|-------|-------|-------|-------|-------|-------|
| $t_1$ | $t_2$ | $t_3$ | $t_4$ | **$t_6$** | **$t_3$** | $t_5$ | $t_6$     | **$t_4$** | **$t_5$** | **$t_2$** | 4 | 4 | 4 | 2 | 2 |
| $t_2$ | **$t_1$** | $t_6$ | $t_5$ | $t_4$ | **$t_6$** | $t_3$ | **$t_4$** | **$t_5$** | $t_3$ | $t_1$ | 2 | 4 | 2 | 2 | 2 |
| $t_3$ | $t_4$ | **$t_1$** | $t_6$ | $t_5$ | $t_1$ | **$t_2$** | **$t_5$** | **$t_6$** | $t_2$ | **$t_4$** | 2 | 4 | 4 | 2 | 2 |
| $t_4$ | **$t_3$** | $t_5$ | **$t_1$** | **$t_2$** | **$t_5$** | $t_6$ | $t_2$ | $t_1$ | **$t_6$** | $t_3$ | 2 | 2 | 4 | 4 | 2 |
| $t_5$ | $t_6$ | **$t_4$** | **$t_2$** | **$t_3$** | $t_4$ | **$t_1$** | $t_3$ | $t_2$ | $t_1$ | **$t_6$** | 2 | 2 | 2 | 4 | 2 |
| $t_6$ | **$t_5$** | **$t_2$** | **$t_3$** | $t_1$ | $t_2$ | **$t_4$** | **$t_1$** | $t_3$ | $t_4$ | $t_5$ | 2 | 2 | 2 | 4 | 4 |

Table 4: Optimal CIRC6 solution, with distance $S + 2(d_2 + d_4) = 14d_1 + 18d_2 + 20d_3 + 18d_4 + 14d_5$.

The solution to CIRC6 (Trick, 2012), where $D_{i,j} = \min\{j - i, 6 - (j - i)\}$ for all $1 \leq i < j \leq 6$, is an element of $S_1$. Table 4 provides this schedule. For each $1 \leq k \leq 5$, we list the number of times the $d_k$ bridge is crossed by each of the six teams.

We conclude the proof by noting that $|S_{i+3}| = |S_i|$ for $2 \leq i \leq 4$, as we can label the teams backward from $t_6$ to $t_1$ to create a feasible schedule where each distance $d_k$ is replaced by $d_{6-k}$. Therefore, we have shown that each $S_i$ is non-empty. □

We are now ready to prove Theorem 1, that the optimal solution to any 6-team instance $\Gamma$ is a schedule that appears in $S_1 \cup S_2 \cup \ldots \cup S_7$. We note that any of these seven optimal distances can be the minimum, depending on the 5-tuple $(d_1, d_2, d_3, d_4, d_5)$.

*Proof.* Suppose the optimal solution to $\Gamma$ has total distance $Z = \sum c_k d_k$. By Lemma 1, $c_1, c_5 \geq 14$, $c_2, c_4 \geq 16$, and $c_3 \geq 20$. Recall that each coefficient $c_k$ is even.

By Lemma 3, $S_1$ is non-empty, and so a schedule cannot be optimal if $Z > S+2(d_2+d_4)$. Thus, if $c_2, c_4 \geq 18$, then we must have $(c_1, c_2, c_3, c_4, c_5) = (14, 18, 20, 18, 14)$ so that $Z = S + 2(d_2 + d_4)$, forcing the schedule to be in set $S_1$.

Suppose that $c_2 \leq c_4$, so that it suffices to check the possibility $c_2 = 16$. By Lemma 2, $t_1$ and $t_2$ must play against each other on Days 1 and 10. There are three cases:

Case 1: $c_2 = 16$, $c_1 = 14$.

Case 2: $c_2 = 16$, $c_1 \geq 16$, $c_4 = 16$.

Case 3: $c_2 = 16$, $c_1 \geq 16$, $c_4 \geq 18$.

In Case 1, every team must travel the minimum number of times across the $d_1$- and $d_2$-bridges: team $t_1$ can only take two road trips, team $t_2$ can only take two road trips to play $\{t_3, t_4, t_5, t_6\}$, and each of $\{t_3, t_4, t_5, t_6\}$ must play their road games against $t_1$ and $t_2$ on consecutive days.

By symmetry, we may assume that the first match between $t_1$ and $t_2$ occurs in the home city of $t_2$ (i.e., it is a road game for $t_1$). By Lemma 2, the schedule for team $t_1$ must be one of the following four cases, for some permutation $\{p, q, r, s\}$ of $\{3, 4, 5, 6\}$.

| Case | Team | 1 | 2 | 3 | 4 | 5 | 6 | 7 | 8 | 9 | 10 |
|---|---|---|---|---|---|---|---|---|---|---|---|
| #A1 | $t_1$ | $t_2$ | $t_?$ | $t_p$ | $t_q$ | $t_?$ | $t_?$ | $t_?$ | $t_r$ | $t_s$ | $t_2$ |
| #A2 | $t_1$ | $t_2$ | $t_?$ | $t_?$ | $t_p$ | $t_q$ | $t_?$ | $t_?$ | $t_r$ | $t_s$ | $t_2$ |
| #A3 | $t_1$ | $t_2$ | $t_?$ | $t_p$ | $t_q$ | $t_r$ | $t_?$ | $t_?$ | $t_?$ | $t_s$ | $t_2$ |
| #A4 | $t_1$ | $t_2$ | $t_?$ | $t_?$ | $t_p$ | $t_q$ | $t_r$ | $t_?$ | $t_?$ | $t_s$ | $t_2$ |

In all four cases, $t_1$ plays a home game against $t_s$ on day 9. In other words, $t_s$ plays on the road against $t_1$ on day 9, forcing $t_s$'s road game against $t_2$ to take place either the day before or the day after. The latter is not possible, as $t_2$ already has a game scheduled against $t_1$ on day 10; thus, $t_s$ must play on the road against $t_2$ on day 8.

Hence, $t_2$ plays a home game against $t_s$ on day 8 and a road game against $t_1$ on day 10. Now suppose that $t_2$ has a home game on day 9. Then $t_2$'s opponent that day must be $t_r$, and we must have either Case #A1 or #A2 above. (This is the only way we can ensure $t_r$ plays their road games against $t_1$ and $t_2$ on consecutive days.)

| Team | 1 | 2 | 3 | 4 | 5 | 6 | 7 | 8 | 9 | 10 |
|---|---|---|---|---|---|---|---|---|---|---|
| $t_1$ | $t_1$ | | | | | | | $t_r$ | $t_s$ | $t_2$ |
| $t_2$ | $t_1$ | | | | | | | $t_s$ | $t_r$ | $t_1$ |

There are six teams in the tournament, and on days 8 and 9, the same set of four teams have each been assigned a game. From the above table, it is clear that teams $t_p$ and $t_q$ must play each other on day 8 and day 9, which is a violation of the *no-repeat* condition. This is a contradiction, and therefore $t_2$ must play a road game on day 9, against some team in $\{t_3, t_4, t_5, t_6\}$.

As mentioned earlier, $t_2$ can only take *two* road trips to play the four teams in $\{t_3, t_4, t_5, t_6\}$, which forces one of the following two scenarios:

| Case | Team | 1 | 2 | 3 | 4 | 5 | 6 | 7 | 8 | 9 | 10 |
|------|------|---|---|---|---|---|---|---|---|---|----|
| #B1 | $t_2$ | $t_1$ | $t_p$ | $t_q$ | $t_?$ | $t_?$ | $t_?$ | $t_r$ | $t_s$ | $t_?$ | $t_1$ |
| #B2 | $t_2$ | $t_1$ | $t_p$ | $t_?$ | $t_?$ | $t_?$ | $t_q$ | $t_r$ | $t_s$ | $t_?$ | $t_1$ |

For each of the $4 \times 2 = 8$ pairs matching the cases for $t_1$ with the cases for $t_2$, we check whether there exists a feasible schedule for which each team in $\{t_3, t_4, t_5, t_6\}$ plays their road games against $t_1$ and $t_2$ on consecutive days. A quick check shows that the *only* possibility is the pairing of Case #A1 with Case #B1, leading to the following schedule for the first two teams:

| Team | 1 | 2 | 3 | 4 | 5 | 6 | 7 | 8 | 9 | 10 |
|------|---|---|---|---|---|---|---|---|---|----|
| $t_1$ | $t_2$ | $t_?$ | $t_p$ | $t_q$ | $t_?$ | $t_?$ | $t_?$ | $t_r$ | $t_s$ | $t_2$ |
| $t_2$ | $t_1$ | $t_p$ | $t_q$ | $t_?$ | $t_?$ | $t_?$ | $t_r$ | $t_s$ | $t_?$ | $t_1$ |

This structural characterization reduces the search space considerably, and from this (see Appendix B) we show that either (i) $c_4 \geq 22$, or (ii) $c_3 \geq 22$ and $c_4 \geq 18$. By Lemma 3, the latter implies $Z = S + 2(d_3 + d_4)$ and the former implies $Z = S + 6d_4$. Therefore, this optimal schedule must be in $S_3$ or $S_4$.

In Case 2, we demonstrate that no structural characterization exists if $c_2 = c_4 = 16$. To do this, we use Lemma 2 (for $c_2 = 16$) and its symmetric analogue (for $c_4 = 16$) to show that in order not to violate the *at-most-three* or *no-repeat* conditions, $t_3$ and $t_4$ must play each other on Days 1 and 10, as well as on some other Day $i$ with $2 \leq i \leq 9$. But then this violates the *each-venue* condition. Hence, we may eliminate this case.

In Case 3, if $c_1 \geq 16$ and $c_4 \geq 18$, then $Z$ is at least $S + 2(d_1 + d_4)$. By Lemma 3, we must have $Z = S + 2(d_1 + d_4)$ and this optimal schedule must be in $S_2$.

So we have shown that if $c_2 = 16$, then the schedule appears in $S_2 \cup S_3 \cup S_4$. By symmetry, if $c_4 = 16$, then the schedule appears in $S_5 \cup S_6 \cup S_7$. Finally, if $c_2, c_4 \geq 18$, the schedule appears in $S_1$. This concludes the proof. □

By Theorem 1, there are only seven possible optimal distances. For each optimal distance, we can enumerate the set of tournament schedules with that distance, thus producing the complete set of possible LD-TTP solutions, over all instances, for the case $n = 6$.

**Theorem 2.** *Consider the set of all feasible tournaments for which the first game between $t_1$ and $t_2$ occurs in the home city of $t_2$. Then there are 295 schedules whose total distance appears in $S_1 \cup S_2 \cup \ldots \cup S_7$, grouped as follows:*

| Total Distance | $\in S_1$ | $\in S_2$ | $\in S_3$ | $\in S_4$ | $\in S_5$ | $\in S_6$ | $\in S_7$ |
|----------------|-----------|-----------|-----------|-----------|-----------|-----------|-----------|
| # of Schedules | 223 | 4 | 8 | 24 | 4 | 8 | 24 |

We derive Theorem 2 by a computer search. For each of $\{S_1, S_2, S_3, S_4\}$, we develop a structural characterization theorem, similar to Case 1 above, that shows that a feasible schedule in that set must have a certain form. This characterization reduces the search space, from which a brute-force search (using Maplesoft) enumerates all possible schedules. While it took several days to enumerate the 223 schedules in $S_1$, Maplesoft took less than 100 seconds to enumerate the set of schedules in $S_3$ and $S_4$. As noted earlier, once we have the set of schedules in $S_i$ (for $2 \leq i \leq 4$), we immediately have the set of schedules in $S_{i+3}$ by symmetry. For the full details of each case, we refer the reader to Appendix B.

Let us briefly explain why $|S_1|$ is odd. For any schedule $S$, let $\Psi(S)$ denote the schedule produced by playing the games backwards (i.e., $t_i$ hosts $t_j$ on day $d$ in $S$ iff $t_i$ hosts $t_j$ on day $(11 - d)$ in $\Psi(S)$.) And let $\Phi(S)$ denote the schedule produced by labelling the six teams in reverse order (i.e., $t_i$ hosts $t_j$ on day $d$ in $S$ iff $t_{7-i}$ hosts $t_{7-j}$ on day $d$ in $\Phi(S)$.) For any schedule $S$, clearly $S \neq \Psi(S)$ and $S \neq \Phi(S)$.

For any schedule $S^* \in S_1$, exactly one of $\Phi(S^*)$ and $\Phi(\Psi(S^*))$ belongs to $S_1$, since we've stipulated that the first game between $t_1$ and $t_2$ occurs in the home city of $t_2$. Since the mapping functions $\Phi$ and $\Phi(\Psi)$ are involutions, the schedules in $S_1$ can be grouped into "pairs". However, in 13 exceptional cases, the schedule $S^* \in S_1$ does not have a pair, since $S^* = \Phi(\Psi(S^*))$. One such example is given in Table 5.

|     | 1     | 2     | 3     | 4     | 5     | 6     | 7     | 8     | 9     | 10    |
| --- | ----- | ----- | ----- | ----- | ----- | ----- | ----- | ----- | ----- | ----- |
| $t_1$ | $t_3$ | $t_6$ | $t_5$ | $\boldsymbol{t_4}$ | $t_3$ | $\boldsymbol{t_5}$ | $t_2$ | $t_4$ | $\boldsymbol{t_6}$ | $\boldsymbol{t_2}$ |
| $t_2$ | $t_4$ | $t_5$ | $\boldsymbol{t_4}$ | $\boldsymbol{t_3}$ | $t_6$ | $t_3$ | $\boldsymbol{t_1}$ | $\boldsymbol{t_6}$ | $t_5$ | $t_1$ |
| $t_3$ | $\boldsymbol{t_1}$ | $\boldsymbol{t_4}$ | $\boldsymbol{t_6}$ | $t_2$ | $t_1$ | $\boldsymbol{t_2}$ | $t_6$ | $t_5$ | $t_4$ | $\boldsymbol{t_5}$ |
| $t_4$ | $\boldsymbol{t_2}$ | $t_3$ | $t_2$ | $t_1$ | $\boldsymbol{t_5}$ | $t_6$ | $t_5$ | $\boldsymbol{t_1}$ | $\boldsymbol{t_3}$ | $\boldsymbol{t_6}$ |
| $t_5$ | $t_6$ | $\boldsymbol{t_2}$ | $\boldsymbol{t_1}$ | $\boldsymbol{t_6}$ | $t_4$ | $t_1$ | $\boldsymbol{t_4}$ | $\boldsymbol{t_3}$ | $t_2$ | $t_3$ |
| $t_6$ | $\boldsymbol{t_5}$ | $\boldsymbol{t_1}$ | $t_3$ | $t_5$ | $\boldsymbol{t_2}$ | $\boldsymbol{t_4}$ | $\boldsymbol{t_3}$ | $t_2$ | $t_1$ | $t_4$ |

Table 5: A schedule $S^*$ in $S_1$ with the property that $S^* = \Phi(\Psi(S^*))$.

In the above schedule, for any pair $(i, j)$, $t_i$ hosts $t_j$ on day $d$ iff $t_{7-i}$ hosts $t_{7-j}$ on day $11 - d$. These thirteen exceptions justify the odd parity of $|S_1|$. For $2 \leq i \leq 7$, there is no schedule with $S^* = \Phi(\Psi(S^*))$, which explains why $|S_i|$ is even in each of these cases.

## 5. An Approximation Algorithm

We have solved the LD-TTP for $n = 4$ and $n = 6$, and in both cases, determined the complete set of schedules attaining the optimal distances. A natural follow-up question is whether our techniques scale for larger values of $n$. To give a partial answer to this question, we show that for all $n \equiv 4 \pmod{6}$, we can develop a solution to the $n$-team LD-TTP whose total distance is at most 33% higher than that of the optimal solution, although in practice this optimality gap is actually much lower.

While our construction is *only* a $\frac{4}{3}$-approximation, we note that this ratio is stronger than the currently best-known $(\frac{5}{3} + \epsilon)$-approximation for the general TTP (Yamaguchi, Imahori, Miyashiro, & Matsui, 2011). Our schedule is based on an "expander construction", and is completely different from previous approaches that generate approximate TTP solutions. We now describe this construction, and apply it to benchmark instances on 10 teams and 16 teams.

Let $m$ be a positive integer. We first create a single round-robin tournament $U$ on $2m$ teams, and then expand this to a double round-robin tournament $T$ on $n = 6m - 2$ teams. We use a variation of the Modified Circle Method (Fujiwara, Imahori, Matsui, & Miyashiro, 2007) to build $U$, our single round-robin schedule. Let $\{u_1, u_2, \ldots, u_{2m-1}, x\}$ be the $2m$ teams. Then each team plays $2m - 1$ games, according to this three-part construction:

(a) For $1 \leq k \leq m$, team $k$ plays the other teams in the following order: $\{2m - k + 1, 2m - k + 2, \ldots, 2m - 1, 1, 2, \ldots, k - 1, x, k + 1, k + 2, \ldots, 2m - k\}$.

(b) For $m + 1 \leq k \leq 2m - 1$, team $k$ plays the other teams in the following order: $\{2m - k + 1, 2m - k + 2, \ldots, k - 1, x, k + 1, k + 2, \ldots, 2m - 1, 1, 2, \ldots, 2m - k\}$.

(c) Team $x$ plays the other teams in the following order: $\{1, m + 1, 2, m + 2, \ldots, m - 1, 2m - 1, m\}$.

|       | 1     | 2     | 3     | 4     | 5     | 6     | 7     |
|-------|-------|-------|-------|-------|-------|-------|-------|
| $u_1$ | $(x)$ | $u_2$ | $u_3$ | $u_4$ | $u_5$ | $u_6$ | $u_7$ |
| $u_2$ | $u_7$ | $u_1$ | $(x)$ | $u_3$ | $u_4$ | $u_5$ | $u_6$ |
| $u_3$ | $u_6$ | $u_7$ | $u_1$ | $u_2$ | $(x)$ | $u_4$ | $u_5$ |
| $u_4$ | $u_5$ | $u_6$ | $u_7$ | $u_1$ | $u_2$ | $u_3$ | $(x)$ |
| $u_5$ | $u_4$ | $(x)$ | $u_6$ | $u_7$ | $u_1$ | $u_2$ | $u_3$ |
| $u_6$ | $u_3$ | $u_4$ | $u_5$ | $(x)$ | $u_7$ | $u_1$ | $u_2$ |
| $u_7$ | $u_2$ | $u_3$ | $u_4$ | $u_5$ | $u_6$ | $(x)$ | $u_1$ |
| $x$   | $u_1$ | $u_5$ | $u_2$ | $u_6$ | $u_3$ | $u_7$ | $u_4$ |

Table 6: The single round-robin construction for $2m = 8$ teams.

For all games not involving team $x$, we designate one home team and one road team as follows: for $1 \leq k \leq m$, $u_k$ plays only road games until it meets team $x$, before finishing the remaining games at home. And for $m + 1 \leq k \leq 2m - 1$, we have the opposite scenario, where $u_k$ plays only home games until it meets team $x$, before finishing the remaining games on the road. As an example, Table 6 provides this single round-robin schedule for the case $m = 4$.

This construction ensures that for any $1 \leq i, j \leq 2m-1$, the match between $u_i$ and $u_j$ has exactly one home team and one road team. To verify this, note that $u_i$ is the home team and $u_j$ is the road team iff $i$ occurs before $j$ in the set $\{1, 2m-1, 2, 2m-2, \ldots, m-1, m+1, m\}$.

Now we "expand" this single round-robin tournament $U$ on $2m$ teams to a double round-robin tournament $T$ on $n = 6m - 2$ teams. To accomplish this, we keep $x$ and transform $u_k$ into three teams, $\{t_{3k-2}, t_{3k-1}, t_{3k}\}$, so that the set of teams in $T$ is precisely $\{t_1, t_2, t_3, \ldots, t_{6m-5}, t_{6m-4}, t_{6m-3}, x\}$.

|            | $6r - 5$    | $6r - 4$    | $6r - 3$    | $6r - 2$    | $6r - 1$    | $6r$        |
|------------|-------------|-------------|-------------|-------------|-------------|-------------|
| $t_{3i-2}$ | $t_{3j-1}$  | $t_{3j}$    | $t_{3j-2}$  | $t_{3j-1}$  | $t_{3j}$    | $t_{3j-2}$  |
| $t_{3i-1}$ | $t_{3j}$    | $t_{3j-2}$  | $t_{3j-1}$  | $t_{3j}$    | $t_{3j-2}$  | $t_{3j-1}$  |
| $t_{3i}$   | $t_{3j-2}$  | $t_{3j-1}$  | $t_{3j}$    | $t_{3j-2}$  | $t_{3j-1}$  | $t_{3j}$    |
| $t_{3j-2}$ | $t_{3i}$    | $t_{3i-1}$  | $t_{3i-2}$  | $t_{3i}$    | $t_{3i-1}$  | $t_{3i-2}$  |
| $t_{3j-1}$ | $t_{3i-2}$  | $t_{3i}$    | $t_{3i-1}$  | $t_{3i-2}$  | $t_{3i}$    | $t_{3i-1}$  |
| $t_{3j}$   | $t_{3i-1}$  | $t_{3i-2}$  | $t_{3i}$    | $t_{3i-1}$  | $t_{3i-2}$  | $t_{3i}$    |

Table 7: Expanding one time slot in $U$ to six time slots in $T$.

Suppose $u_i$ is the home team in its game against $u_j$, played in time slot $r$. Then we expand that time slot in $U$ into *six* time slots in $T$, namely the slots $6r - 5$ to $6r$. We describe the match assignments in Table 7.

Before proceeding further, let us explain the idea behind this construction. Recall that by the *each-venue* condition, each team in $T$ must visit every opponent's home stadium exactly once, and by the *at-most-three* condition, road trips are at most three games. We will build a tournament that maximizes the number of three-game road trips, and ensure that the majority of these road trips involve three venues closely situated to one another, to minimize total travel. In Table 7 above, if $\{t_{3j-2}, t_{3j-1}, t_{3j}\}$ are located in the same region, then each of the teams in $\{t_{3i-2}, t_{3i-1}, t_{3i}\}$ can play their three road games against these teams in a highly-efficient manner.

We now explain how to expand the time slots in games involving team $x$. For each $1 \leq k \leq m$, consider the game between $u_k$ and $x$. We expand that time slot in $U$ into six time slots in $T$, as described in Table 8.

|           | $6r-5$      | $6r-4$      | $6r-3$      | $6r-2$      | $6r-1$      | $6r$        |
|-----------|-------------|-------------|-------------|-------------|-------------|-------------|
| $t_{3k-2}$| **$x$**     | **$t_{3k}$**| $t_{3k-1}$  | $x$         | $t_{3k}$    | **$t_{3k-1}$**|
| $t_{3k-1}$| $t_{3k}$    | **$x$**     | **$t_{3k-2}$**| $t_{3k}$  | $x$         | $t_{3k-2}$  |
| $t_{3k}$  | **$t_{3k-1}$**| $t_{3k-2}$| $x$         | $t_{3k-1}$  | **$t_{3k-2}$**| **$x$**   |
| $x$       | $t_{3k-2}$  | $t_{3k-1}$  | **$t_{3k}$**| **$t_{3k-2}$**| **$t_{3k-1}$**| $t_{3k}$|

Table 8: The six time slot expansion for $1 \leq k \leq m$.

And for each $m + 1 \leq k \leq 2m - 1$, consider the game between $u_k$ and $x$. We expand that time slot in $U$ into six time slots in $T$, as described in Table 9.

|           | $6r-5$      | $6r-4$      | $6r-3$      | $6r-2$      | $6r-1$      | $6r$        |
|-----------|-------------|-------------|-------------|-------------|-------------|-------------|
| $t_{3k-2}$| $x$         | **$t_{3k}$**| **$t_{3k-1}$**| $x$       | $t_{3k}$    | $t_{3k-1}$  |
| $t_{3k-1}$| **$t_{3k}$**| $x$         | $t_{3k-2}$  | $t_{3k}$    | **$x$**     | **$t_{3k-2}$**|
| $t_{3k}$  | $t_{3k-1}$  | $t_{3k-2}$  | **$x$**     | **$t_{3k-1}$**| **$t_{3k-2}$**| $x$       |
| $x$       | **$t_{3k-2}$**| **$t_{3k-1}$**| $t_{3k}$| $t_{3k-2}$  | $t_{3k-1}$  | **$t_{3k}$**|

Table 9: The six time slot expansion for $m + 1 \leq k \leq 2m - 1$.

This construction builds a double round-robin tournament $T$ with $n = 6m - 2$ teams and $2n - 2 = 12m - 6$ time slots. To give an example, Table 10 provides $T$ for the case $m = 2$.

It is straightforward to verify that this tournament schedule on $n = 6m - 2$ teams is feasible for all $m \geq 1$, i.e., it satisfies the *each-venue*, *at-most-three*, and *no-repeat* conditions. We now show that this expander construction gives a $\frac{4}{3}$-approximation to the LD-TTP, regardless of the values of the distance parameters $d_1, d_2, \ldots, d_{n-1}$.

Let $\Gamma$ be an $n$-team instance of the LD-TTP, with $n = 6m - 2$ for some $m \geq 1$. Let $S$ be the total distance of the optimal solution of $\Gamma$. Using our expander construction, we generate a feasible tournament with total distance less than $\frac{4}{3}S$. This gives a $\frac{4}{3}$-approximation to the LD-TTP.

|   | 1 | 2 | 3 | 4 | 5 | 6 | 7 | 8 | 9 | 10 | 11 | 12 | 13 | 14 | 15 | 16 | 17 | 18 |
|---|---|---|---|---|---|---|---|---|---|----|----|----|----|----|----|----|----|----|
| $t_1$ | $x$ | $t_3$ | $t_2$ | $x$ | $t_3$ | $t_2$ | $t_5$ | $t_6$ | $t_4$ | $t_5$ | $t_6$ | $t_4$ | $t_8$ | $t_9$ | $t_7$ | $t_8$ | $t_9$ | $t_7$ |
| $t_2$ | $t_3$ | $x$ | $t_1$ | $t_3$ | $x$ | $t_1$ | $t_6$ | $t_4$ | $t_5$ | $t_6$ | $t_4$ | $t_5$ | $t_9$ | $t_7$ | $t_8$ | $t_9$ | $t_7$ | $t_8$ |
| $t_3$ | $t_2$ | $t_1$ | $x$ | $t_2$ | $t_1$ | $x$ | $t_4$ | $t_5$ | $t_6$ | $t_4$ | $t_5$ | $t_6$ | $t_7$ | $t_8$ | $t_9$ | $t_7$ | $t_8$ | $t_9$ |
| $t_4$ | $t_9$ | $t_8$ | $t_7$ | $t_9$ | $t_8$ | $t_7$ | $t_3$ | $t_2$ | $t_1$ | $t_3$ | $t_2$ | $t_1$ | $x$ | $t_6$ | $t_5$ | $x$ | $t_6$ | $t_5$ |
| $t_5$ | $t_7$ | $t_9$ | $t_8$ | $t_7$ | $t_9$ | $t_8$ | $t_1$ | $t_3$ | $t_2$ | $t_1$ | $t_3$ | $t_2$ | $t_6$ | $x$ | $t_4$ | $t_6$ | $x$ | $t_4$ |
| $t_6$ | $t_8$ | $t_7$ | $t_9$ | $t_8$ | $t_7$ | $t_9$ | $t_2$ | $t_1$ | $t_3$ | $t_2$ | $t_1$ | $t_3$ | $t_5$ | $t_4$ | $x$ | $t_5$ | $t_4$ | $x$ |
| $t_7$ | $t_5$ | $t_6$ | $t_4$ | $t_5$ | $t_6$ | $t_4$ | $x$ | $t_9$ | $t_8$ | $x$ | $t_9$ | $t_8$ | $t_3$ | $t_2$ | $t_1$ | $t_3$ | $t_2$ | $t_1$ |
| $t_8$ | $t_6$ | $t_4$ | $t_5$ | $t_6$ | $t_4$ | $t_5$ | $t_9$ | $x$ | $t_7$ | $t_9$ | $x$ | $t_7$ | $t_1$ | $t_3$ | $t_2$ | $t_1$ | $t_3$ | $t_2$ |
| $t_9$ | $t_4$ | $t_5$ | $t_6$ | $t_4$ | $t_5$ | $t_6$ | $t_8$ | $t_7$ | $x$ | $t_8$ | $t_7$ | $x$ | $t_2$ | $t_1$ | $t_3$ | $t_2$ | $t_1$ | $t_3$ |
| $x$ | $t_1$ | $t_2$ | $t_3$ | $t_1$ | $t_2$ | $t_3$ | $t_7$ | $t_8$ | $t_9$ | $t_7$ | $t_8$ | $t_9$ | $t_4$ | $t_5$ | $t_6$ | $t_4$ | $t_5$ | $t_6$ |

Table 10: The case $m = 2$, producing a 10-team tournament.

Let $y_1, y_2, \ldots, y_n$ be the $n = 6m - 2$ teams of $\Gamma$, in that order, with $d_k$ being the distance from $y_k$ to $y_{k+1}$ for all $1 \le k \le n - 1$. Now we map the set $\{t_1, t_2, \ldots, t_{n-1}, x\}$ to $\{y_1, y_2, \ldots, y_n\}$ as follows: $t_i = y_i$ for $1 \le i \le 3m - 3$, $x = y_{3m-2}$, and $t_i = y_{i+1}$ for $3m - 2 \le i \le 6m - 3$. In Figure 4 below, we illustrate this mapping for the case $m = 2$, where the $n = 6m - 2$ teams are divided into three triplets and a singleton $x$:

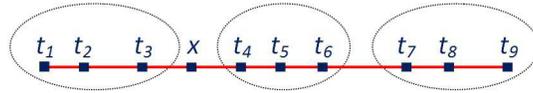

Figure 4: The labeling of the $n = 6m - 2$ teams, for $m = 2$.

We then apply this labeling to our expander construction to create a feasible $n$-team tournament $T$, where $n = 6m - 2$ for some $m \ge 1$. The following theorem tells us the total distance of this tournament, as a function of the $n - 1$ distance parameters $d_1, d_2, \ldots, d_{n-1}$.

**Theorem 3.** *Let $T$ be the $n$-team double round-robin tournament created by our expander construction, where $n = 6m - 2$. For each $1 \le k \le 6m - 3$, let $f_k$ be the total number of times the $d_k$-length "bridge" is crossed, so that the total distance of $T$ is $\sum_{k=1}^{n-1} f_k d_k$. Then the value of $f_k$ is given by Table 11. In addition, $f_1 = (8n - 8)/3$, $f_2 = 4n - 4$, $f_{3m-2} = f_{n/2-1} = (n^2 + 6n - 16)/3$, $f_{3m-1} = f_{n/2} = (n^2 + 9n - 22)/3$, $f_{3m} = f_{n/2+1} = (n^2 + 9n - 34)/3$, and $f_{6m-3} = f_{n-1} = (8n - 2)/3$.*

| Case | $k$ | $f_k$ |
|---|---|---|
| (a) | $k = 4, 7, 10, \ldots, 3m - 5$ | $4k(n - k)/3 + (6n + 8k - 20)/3$ |
| (b) | $k = 5, 8, 11, \ldots, 3m - 4$ | $4k(n - k)/3 + (4n + 12k - 20)/3$ |
| (c) | $k = 3, 6, 9, 12, \ldots, 3m - 3$ | $4k(n - k)/3 + (4n + 6k - 16)/3$ |
| (d) | $k = 3m + 1, 3m + 4, \ldots, 6m - 8, 6m - 5$ | $4k(n - k)/3 + (8n - 4k - 22)/3$ |
| (e) | $k = 3m + 2, 3m + 5, \ldots, 6m - 7, 6m - 4$ | $4k(n - k)/3 + (14n - 10k - 16)/3$ |
| (f) | $k = 3m + 3, 3m + 6, \ldots, 6m - 6$ | $4k(n - k)/3 + (4n - 2k - 4)$ |

Table 11: The formulas for $f_k$ as a function of $n$ and $k$.

*Proof.* For each of the six cases, we carefully enumerate the number of times each team crosses the bridge, by considering the activity of each team in the tournament schedule $T$.

(a) Of the $k$ teams to the left of the $d_k$-length bridge, one team crosses the bridge $2(n-k)/3$ times, $(k+5)/3$ teams cross the bridge $2(n-k+3)/3$ times and $(2k-8)/3$ teams cross the bridge $2(n-k+6)/3$ times. And of the $n-k-1$ teams to the right of the bridge (not including team $x$), $(2n-3k-5)/3$ of these teams cross the bridge $2(k+2)/3$ times and the remaining $(n+2)/3$ teams cross the bridge $2(k+5)/3$ times. Finally, team $x$ crosses the bridge $(4k+2)/3$ times. From there, we sum up the cases and determine that $f_k = 4k(n-k)/3 + (6n+8k-20)/3$.

(b) Of the $k$ teams to the left of the $d_k$-length bridge, one team crosses the bridge $2(n-k+1)/3$ times, $(k+4)/3$ teams cross the bridge $2(n-k+4)/3$ times and $(2k-7)/3$ teams cross the bridge $2(n-k+7)/3$ times. And of the $n-k-1$ teams to the right of the bridge (not including team $x$), $(2n-3k-2)/3$ of these teams cross the bridge $2(k+1)/3$ times, $(n-4)/3$ teams cross the bridge $2(k+4)/3$ times, and one team crosses $2(k+7)/3$ times. Finally, team $x$ crosses the bridge $(4k-2)/3$ times. From there, we sum up the cases and determine that $f_k = 4k(n-k)/3 + (4n+12k-20)/3$.

(c) Of the $k$ teams to the left of the $d_k$-length bridge, $(k+6)/3$ of these teams cross the bridge $2(n-k+2)/3$ times, and the remaining $(2k-6)/3$ teams cross the bridge $2(n-k+5)/3$ times. And of the $n-k-1$ teams to the right of the bridge (not including team $x$), $(n-k-1)/3$ of these teams cross the bridge $2k/3$ times and the remaining $2(n-k-1)/3$ teams cross the bridge $(2k+6)/3$ times. Finally, team $x$ crosses the bridge $4k/3$ times. From there, we sum up the cases and determine that $f_k = 4k(n-k)/3 + (4n+6k-16)/3$.

(d) Of the $k-1$ teams to the left of the $d_k$-length bridge (not including team $x$), $(k+5)/3$ teams cross the bridge $2(n-k)/3$ times, and the remaining $(2k-8)/3$ teams cross the bridge $2(n-k+3)/3$ times. And of the $n-k$ teams to the right of the bridge, $(n-k+3)/3$ cross the bridge $2(k+2)/3$ times and the remaining $(2n-2k-3)/3$ teams cross the bridge $2(k+5)/3$ times. Finally, team $x$ crosses the bridge $2(n-k)/3$ times. From there, we sum up the cases and determine that $f_k = 4k(n-k)/3 + (8n-4k-22)/3$.

(e) Of the $k-1$ teams to the left of the $d_k$-length bridge (not including team $x$), $(3k-n+4)/3$ teams cross the bridge $2(n-k+1)/3$ times, and the remaining $(n-7)/3$ teams cross the bridge $2(n-k+4)/3$ times. And of the $n-k$ teams to the right of the bridge, $(n-k+4)/3$ cross the bridge $2(k+4)/3$ times and the remaining $(2n-2k-4)/3$ teams cross the bridge $2(k+7)/3$ times. Finally, team $x$ crosses the bridge $2(n-k+4)/3$ times. From there, we sum up the cases and determine that $f_k = 4k(n-k)/3 + (14n-10k-16)/3$.

(f) Of the $k-1$ teams to the left of the $d_k$-length bridge (not including team $x$), $(3k-n+1)/3$ teams cross the bridge $2(n-k+2)/3$ times, and the remaining $(n-4)/3$ teams cross the bridge $2(n-k+5)/3$ times. And of the $n-k$ teams to the right of the bridge, $(n-k+2)/3$ cross the bridge $2(k+3)/3$ times and the remaining $(2n-2k-2)/3$ teams cross the bridge $2(k+6)/3$ times. Finally, team $x$ crosses the bridge $2(n - k + 2)/3$ times. From there, we sum up the cases and determine that $f_k = 4k(n - k)/3 + (4n - 2k - 4)$.

Finally, we clear all the exceptional cases. If $k = 1$, team $t_1$ crosses the bridge $2(n-1)/3$ times, while the remaining $n - 1$ teams cross twice. Thus, $f_1 = 2(n - 1)/3 + 2(n - 1) = (8n - 8)/3$. If $k = 2$, team $t_1$ crosses the bridge $2(n-1)/3$ times, team $t_2$ crosses $2(n+2)/3$ times, $(2n - 5)/3$ teams cross twice, and $(n - 1)/3$ teams cross four times. Thus, $f_2 = 2(n - 1)/3 + 2(n + 2)/3 + (4n - 10)/3 + (4n - 4)/3 = 4n - 4$. If $k = n - 1$, team $t_n$ crosses the bridge $2(n + 2)/3$ times, while the remaining $n - 1$ teams cross twice. Thus, $f_{n-1} = 2(n + 2)/3 + 2(n - 1) = (8n - 2)/3$.

For $k = \frac{n}{2} - 1$, the formula for $f_k$ is the same as that of case (d), except that one team makes an additional trip across the bridge. For $k = \frac{n}{2} - 1$, the formula for $f_k$ is the same as that of case (e), except that one team makes one fewer trip across the bridge. Finally, for $k = \frac{n}{2} + 1$, the formula for $f_k$ is the same as that of case (f), except that two teams make one additional trip across the bridge. A straightforward calculation then results in verifying that $f_{3m-2} = f_{n/2-1} = (n^2 + 6n - 16)/3$, $f_{3m-1} = f_{n/2} = (n^2 + 9n - 22)/3$, and $f_{3m} = f_{n/2+1} = (n^2 + 9n - 34)/3$. This completes the proof. □

For example, for the case $m = 2$ (see Table 10), we have $n = 10$, and so the total travel distance of $T$ is $24d_1 + 36d_2 + 42d_3 + 48d_4 + 56d_5 + 52d_6 + 38d_7 + 36d_8 + 26d_9$.

Let $S = \sum_{k=1}^{n-1} l_k d_k$ be the trivial lower bound of $\Gamma$, found by adding the independent lower bounds for each team $t_i$. As we described in the proof of Lemma 1, we have $l_k = 2k\lceil\frac{n-k}{3}\rceil + 2(n - k)\lceil\frac{k}{3}\rceil$ because each of the $k$ teams to the left of the $d_k$ bridge must make at least $2\lceil\frac{n-k}{3}\rceil$ trips across the bridge, and the $n - k$ teams to the right of this bridge must make at least $2\lceil\frac{k}{3}\rceil$ trips across.

For $m \geq 3$, it is straightforward to verify that $\frac{f_k}{l_k} < \frac{4}{3}$ for all $1 \leq k \leq n - 1$, thus establishing our $\frac{4}{3}$-approximation for the LD-TTP. This ratio of $\frac{4}{3}$ is the best possible due to the case $k = 3$, which has $f_3 = \frac{16n-34}{3}$ and $l_3 = 4n - 8$, implying $\frac{f_3}{l_3} \to \frac{4}{3}$ as $n \to \infty$. This worst-case scenario is achieved when $d_k = 0$ for all $k \neq 3$, i.e., when teams $\{t_1, t_2, t_3\}$ are located at one vertex, and the remaining $n - 3$ teams are located at another vertex.

A natural question is whether there exist similar constructions for $n \equiv 0$ and $n \equiv 2$ (mod 6). In these cases, in addition to the $n \equiv 4$ case we just analyzed, we ask whether a $\frac{4}{3}$-approximation is best possible. This is just one of many open questions arising from this work.

## 6. Application to Benchmark Sets

We now apply our theories to various benchmark TTP sets. We start with the case $n = 6$, and apply Theorems 1 and 2 to all known 6-team TTP benchmarks. In addition to NL6, we examine a six-team set from the Super Rugby League (SUPER6), six galaxy stars whose coordinates appear in three-dimensional space (GALAXY6), our earlier six-team circular distance instance (CIRC6), and the trivial constant distance instance (CON6) where each pair of teams has a distance of one unit.

For all our benchmark sets, we first order the six teams so that they approximate a straight line, either through a formal "line of best fit" or an informal check by inspection.

Having ordered our six teams, we determine the five-tuple $(d_1, d_2, d_3, d_4, d_5)$ from the distance matrix and ignore the other $\binom{6}{2} - 5 = 10$ entries. Modifying our benchmark set and assuming the six teams lie on a straight line, we solve the LD-TTP via Theorem 1. Using Theorem 2, we take the set of tournament schedules achieving this optimal distance and apply the actual distance matrix of the benchmark set (with all $\binom{6}{2}$ entries) to each of these optimal schedules and output the tournament with the minimum total distance.

This simple process, each taking approximately 0.3 seconds of computation time per benchmark set, generates a feasible solution to the 6-team TTP. To our surprise, this algorithm outputs the distance-optimal schedule in all five of our benchmark sets. This was an unexpected result, given the non-linearity of our data sets: for example, CIRC6 has the teams arranged in a circle, while GALAXY6 uses three-dimensional distances. To illustrate our theory, let us begin with NL6, ordering the six teams from south to north:

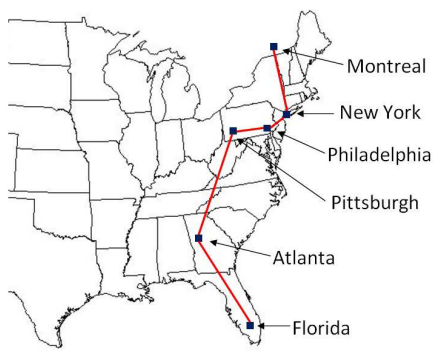

Figure 5: Location of the six NL6 teams.

Thus, Florida is $t_1$, Atlanta is $t_2$, Pittsburgh is $t_3$, Philadelphia is $t_4$, New York is $t_5$, and Montreal is $t_6$. From the NL6 distance matrix (Trick, 2012), we have $(d_1, d_2, d_3, d_4, d_5) = (605, 521, 257, 80, 337)$.

Since $2\min\{d_2 + d_4, d_1 + d_4, d_3 + d_4, 3d_4, d_2 + d_5, d_2 + d_3, 3d_2\} = 6d_4 = 480$, Theorem 1 tells us that the optimal LD-TTP solution has total distance $S + 6d_4 = 14d_1 + 16d_2 + 20d_3 + 22d_4 + 14d_5 = 28424$. By Theorem 2, there are 24 schedules in set $S_4$, all with total distance $S + 6d_4$. Two of these 24 schedules are presented in Table 12.

|       | 1     | 2     | 3     | 4     | 5     | 6     | 7     | 8     | 9     | 10    |
|-------|-------|-------|-------|-------|-------|-------|-------|-------|-------|-------|
| $t_1$ | $t_2$ | $t_4$ | **$t_5$** | **$t_3$** | $t_5$ | $t_6$ | $t_3$ | **$t_4$** | **$t_6$** | **$t_2$** |
| $t_2$ | **$t_1$** | **$t_5$** | **$t_3$** | $t_4$ | $t_6$ | $t_3$ | **$t_4$** | **$t_6$** | $t_5$ | $t_1$ |
| $t_3$ | **$t_5$** | **$t_6$** | $t_2$ | $t_1$ | **$t_4$** | **$t_2$** | **$t_1$** | $t_5$ | $t_4$ | $t_6$ |
| $t_4$ | **$t_6$** | **$t_1$** | **$t_6$** | **$t_2$** | $t_3$ | **$t_5$** | $t_2$ | $t_1$ | **$t_3$** | $t_5$ |
| $t_5$ | $t_3$ | $t_2$ | $t_1$ | **$t_6$** | **$t_1$** | $t_4$ | $t_6$ | **$t_3$** | **$t_2$** | **$t_4$** |
| $t_6$ | **$t_4$** | $t_3$ | $t_4$ | $t_5$ | **$t_2$** | **$t_1$** | **$t_5$** | $t_2$ | $t_1$ | **$t_3$** |

|       | 1     | 2     | 3     | 4     | 5     | 6     | 7     | 8     | 9     | 10    |
|-------|-------|-------|-------|-------|-------|-------|-------|-------|-------|-------|
| $t_1$ | $t_2$ | $t_5$ | **$t_6$** | **$t_3$** | $t_6$ | $t_4$ | $t_3$ | **$t_5$** | **$t_4$** | **$t_2$** |
| $t_2$ | **$t_1$** | **$t_6$** | **$t_3$** | $t_5$ | $t_4$ | $t_3$ | **$t_5$** | **$t_4$** | $t_6$ | $t_1$ |
| $t_3$ | **$t_6$** | $t_4$ | $t_2$ | $t_1$ | **$t_5$** | **$t_2$** | **$t_1$** | $t_6$ | $t_5$ | **$t_4$** |
| $t_4$ | **$t_5$** | **$t_3$** | $t_5$ | $t_6$ | **$t_2$** | **$t_1$** | **$t_6$** | $t_2$ | $t_1$ | $t_3$ |
| $t_5$ | $t_4$ | **$t_1$** | $t_4$ | **$t_2$** | $t_3$ | **$t_6$** | $t_2$ | $t_1$ | **$t_3$** | $t_6$ |
| $t_6$ | $t_3$ | $t_2$ | $t_1$ | **$t_4$** | **$t_1$** | $t_5$ | $t_4$ | **$t_3$** | **$t_2$** | **$t_5$** |

Table 12: Two LD-TTP solutions with total distance $S + 6d_4$.

Removing this straight line assumption, we now apply the actual NL6 distance matrix to determine the total distance traveled for each of these 24 schedules from set $S_4$, which will naturally produce different sums. The left schedule in Table 12 is best among the 24 schedules, with total distance 23916, while the right schedule is the worst, with total distance 24530. We note that the left schedule, achieving the optimal distance of 23916 miles, is identical to Table 1.

We repeat the same analysis with the other four benchmark sets. In each, we mark which of the sets $\{S_1, S_2, \ldots, S_7\}$ produced the optimal schedule.

| Benchmark Data Set | Optimal Solution | LD-TTP Solution | Optimal Schedule |
|---|---|---|---|
| NL6 | 23916 | 23916 | $\in S_4$ |
| SUPER6 | 130365 | 130365 | $\in S_3$ |
| GALAXY6 | 1365 | 1365 | $\in S_1$ |
| CIRC6 | 64 | 64 | $\in S_1$ |
| CON6 | 43 | 43 | $\in S_1$ |

Table 13: Comparing LD-TTP to TTP on benchmark data sets.

A sophisticated branch-and-price heuristic (Irnich, 2010) solved NL6 in just over one minute, yet required three hours to solve CIRC6. The latter problem was considerably more difficult due to the inherent symmetry of the data set, which required more branching. However, through our LD-TTP approach, both problems can be solved to optimality in the same amount of time – approximately 0.3 seconds.

Based on the results of Table 13, we ask whether there exists a 6-team instance $\Gamma$ where the optimal TTP solution is different from the optimal LD-TTP solution. This question will be answered in the following section.

To conclude this section, we apply the $\frac{4}{3}$-approximation produced by our expander construction to various (non-linear) benchmark sets with $n \equiv 4 \pmod{6}$. We apply our construction to the 10-team and 16-team instances of our earlier examples (Trick, 2012).

| Instance | Optimal | Our Solution | Percentage Gap |
|---|---|---|---|
| CONS10 | 124 | 128 | 3.2% |
| CIRC10 | 242 | 276 | 14.0% |
| NL10 | 59436 | 63850 | 7.4% |
| SUPER10 | 316329 | 361924 | 14.4% |
| GALAXY10 | 4535 | 4862 | 7.2% |
| CONS16 | 327 | 334 | 2.1% |
| CIRC16 | $[846, 916]$ | 994 | $[8.5\%, 17.5\%]$ |
| NL16 | $[249477, 261687]$ | 286439 | $[9.5\%, 14.8\%]$ |
| GALAXY16 | $[13619, 14900]$ | 15429 | $[3.6\%, 13.3\%]$ |

Table 14: Comparing our construction to the optimal solution in nine benchmark sets.

For the GALAXY, NL, and SUPER instances, we first need to arrange the $n$ teams to approximate a straight line. To do this, we apply a simple algorithm that first randomly assigns the $n$ teams to $\{t_1, t_2, \ldots, t_{n-1}, x\}$, and calculates the sum total of distances between each "adjacent" pair of teams. We generate a local line-of-best-fit by recursively selecting two teams $t_i$ and $t_j$ and switching their positions if it reduces the sum of these $n-1$ distances. The algorithm terminates with a permutation of the $n$ teams to $\{t_1, t_2, \ldots, t_{n-1}, x\}$ that is locally optimal (but perhaps not globally), from which we apply the expander construction to calculate the total travel distance of our $n$-team tournament.

Instead of a time-consuming process that enumerates all $n!$ permutations of the teams, our simple algorithm generates a fast solution to each of our benchmark instances in less than 2 seconds of total computation time. Despite the simplicity of our approach, we see in Table 14 that the optimality gap is extremely small for the constant instances (CONS), and is quite reasonable for all the other (non-linear) instances.

## 7. Optimality Gap

In Table 13, all five of the 6-team benchmark instances produced identical solutions for both the TTP and LD-TTP. A natural question is whether this is always the case. We show that the TTP and LD-TTP solutions must be identical for $n = 4$, but not necessarily for $n = 6$.

For any instance $\Gamma$ on $n$ teams, define $X_\Gamma$ to be the total distance of an optimal TTP solution, and $X_\Gamma^*$ to be the total distance of an optimal LD-TTP solution. Define $OG_n$ to be the *maximum optimality gap*, the largest value of $\frac{X_\Gamma^* - X_\Gamma}{X_\Gamma}$ taken over all instances $\Gamma$.

**Theorem 4.** *For any instance $\Gamma$ on $n = 4$ teams, the optimal TTP solution is the optimal LD-TTP solution. In other words, $OG_4 = 0\%$.*

*Proof.* In Table 3, we showed that there are 18 non-isomorphic schedules with total distance $8(d_1 + d_2 + d_3)$, i.e., 18 different solutions to the LD-TTP. For each of these 18 schedules, we remove the linear distance assumption and determine the total travel distance as a function of the six distance parameters (i.e., the variables in $\{D_{i,j} : 1 \leq i < j \leq 4\}$). For example, the schedule in Table 2 has total distance $4D_{1,2} + 2D_{1,3} + 2D_{1,4} + 3D_{2,3} + D_{2,4} + 5D_{3,4}$ which we represent by the 6-tuple $(4, 2, 2, 3, 1, 5)$. Considering all $4!$ permutations of $\{t_1, t_2, t_3, t_4\}$, there are $18 \times 24$ tournament schedules, producing 36 unique 6-tuples, including $(4, 2, 2, 3, 1, 5)$. Denote by $L$ this set of thirty-six 6-tuples.

A brute-force enumeration finds 1920 feasible 4-team tournaments. For each of these 1920 tournaments, we determine the 6-tuple representing the total travel distance, and find 246 unique 6-tuples, which we denote by set $A$. By definition, $L \subset A$.

To prove that $OG_4 = 0$, we must verify that for any set $\{D_{1,2}, D_{1,3}, D_{1,4}, D_{2,3}, D_{2,4}, D_{3,4}\}$ satisfying the Triangle Inequality, the optimal solutions of the TTP and LD-TTP are the same, i.e., the optimal solution among all schedules (whose six-tuples are given by $A$) appears in the subset of linear-distance schedules (whose six-tuples are given by $L$). To establish this, we first use the Triangle Inequality to verify that for 204 of the $246 - 36 = 210$ elements in $A \backslash L$, the corresponding schedule is dominated by at least one of the elements in $L$.

For example, the six-tuple $(3, 4, 3, 4, 1, 4)$ is one of the 210 elements in $A \backslash L$. Comparing this with the six-tuple $(4, 2, 2, 3, 1, 5) \in L$, we see that the corresponding schedule in $A \backslash L$ has total distance $2D_{1,3} + D_{1,4} + D_{2,3} - D_{1,2} - D_{3,4} = (D_{1,3} + D_{1,4} - D_{3,4}) + (D_{1,3} + D_{2,3} - D_{1,2}) \geq 0$ more than the corresponding schedule in $L$, which is given in Table 2.

A computer search shows that 204 of the 210 elements in $A \backslash L$ can be handled by applying the Triangle Inequality in this way, showing it is dominated by at least one element in $L$. There are just six "exceptions", namely the 6-tuples in the set $\{(2, 3, 3, 3, 3, 4), (3, 2, 3, 3, 4, 3), (3, 3, 2, 4, 3, 3), (3, 3, 4, 2, 3, 3), (3, 4, 3, 3, 2, 3),$ and $(4, 3, 3, 3, 3, 2)\}$. In these cases, the analysis is slightly harder. Consider the six-tuple $(2, 3, 3, 3, 3, 4)$; the rest can be handled in the same way, by symmetry.

There are twelve 6-tuples in $L$ which have 17 total trips, where the $D_{1,2}$ coefficient is strictly less than the $D_{4,5}$ coefficient. (An example of one such 6-tuple is $(4, 2, 2, 3, 1, 5)$.) Taking its average, we derive the 6-tuple $(7/3, 17/6, 17/6, 17/6, 17/6, 10/3)$, implying the existence of at least one LD-TTP schedule whose total distance $X$ is at most $(14D_{1,2} + 17D_{1,3} + 17D_{1,4} + 17D_{2,3} + 17D_{2,4} + 20D_{3,4})/6$. Let $Y$ be the distance represented by the 6-tuple $(2, 3, 3, 3, 3, 4)$. Then by the Triangle Inequality,

$$\begin{aligned} 6(Y - X) &= -2D_{1,2} + D_{1,3} + D_{1,4} + D_{2,3} + D_{2,4} + 4D_{3,4} \\ &= (D_{1,3} + D_{3,4} + D_{4,2} - D_{2,1}) + (D_{1,4} + D_{4,3} + D_{3,2} - D_{2,1}) + 2D_{3,4} \\ &\geq 0 + 0 + 2 \times 0 = 0. \end{aligned}$$

In other words, we have shown that every element in $A \backslash L$ is dominated by at least one element in $L$. Therefore, for any choice of $\{D_{1,2}, D_{1,3}, D_{1,4}, D_{2,3}, D_{2,4}, D_{3,4}\}$ satisfying the Triangle Inequality, the optimal solutions of the TTP and LD-TTP are the same, i.e., $OG_4 = 0\%$. □

In an earlier paper (Hoshino & Kawarabayashi, 2012), the authors conjectured that $OG_6 > 0\%$, although we were unable to find a 6-team instance with a positive optimality gap. Here we present a simple instance to show that $OG_6 \geq \frac{1}{43} \sim 2.3\%$.

Let $\Gamma$ be the 6-team instance with $D_{1,2} = D_{5,6} = 2$ and all other $D_{i,j} = 1$. Clearly these $\binom{6}{2} = 15$ distances satisfy the Triangle Inequality. We now show that $X_\Gamma = 43$ and $X_\Gamma^* = 44$, thus proving that $OG_6 \geq \frac{1}{43}$. Consider Table 15, which is a solution to the TTP (but not LD-TTP) with 43 trips.

|       | 1     | 2     | 3     | 4     | 5     | 6     | 7     | 8     | 9     | 10    | # of Trips |
|-------|-------|-------|-------|-------|-------|-------|-------|-------|-------|-------|------------|
| $t_1$ | **$t_5$** | **$t_2$** | **$t_3$** | $t_5$ | $t_3$ | **$t_4$** | **$t_6$** | $t_4$ | $t_2$ | $t_6$ | 7 |
| $t_2$ | $t_6$ | $t_1$ | $t_5$ | **$t_6$** | **$t_5$** | $t_3$ | $t_4$ | $t_3$ | **$t_1$** | **$t_4$** | 7 |
| $t_3$ | $t_4$ | $t_6$ | $t_1$ | **$t_4$** | **$t_1$** | $t_2$ | **$t_5$** | **$t_2$** | **$t_6$** | $t_5$ | 8 |
| $t_4$ | **$t_3$** | **$t_5$** | **$t_6$** | $t_3$ | $t_6$ | $t_1$ | **$t_2$** | **$t_1$** | $t_5$ | $t_2$ | 7 |
| $t_5$ | $t_1$ | $t_4$ | **$t_2$** | **$t_1$** | $t_2$ | $t_6$ | $t_3$ | **$t_6$** | **$t_4$** | **$t_3$** | 7 |
| $t_6$ | **$t_2$** | **$t_3$** | $t_4$ | $t_2$ | **$t_4$** | **$t_5$** | $t_1$ | $t_5$ | $t_3$ | **$t_1$** | 7 |

Table 15: An optimal 43-trip TTP solution that beats the optimal LD-TTP solution.

By inspection, we see that no team travels along the bridge connecting the stadiums of $t_1$ and $t_2$, or along the bridge connecting the stadiums of $t_5$ and $t_6$. Thus, the total travel distance must be $43 \times 1 = 43$, since the 2-unit distances $D_{1,2}$ and $D_{5,6}$ do not appear in the total sum. Since every 6-team tournament must have at least 43 total trips (see Table 13), this proves that $X_\Gamma = 43$.

For each of the 295 potentially-optimal LD-TTP schedules in Theorem 2, we consider all $6! = 720$ permutations of $(t_1, t_2, t_3, t_4, t_5, t_6)$ to see if any tournament can have total distance 43. A computer search shows that 36 of the 295 schedules can have total distance 44, but none can have distance 43. This proves that $X_\Gamma^* = 44$ is the optimal LD-TTP travel distance for this instance $\Gamma$.

Therefore, the maximum optimality gap $OG_6$ is at least $\frac{1}{43} \sim 2.3\%$. We ask whether this gap can be made larger, and propose the following question.

**Problem 1.** *Determine the value of $OG_n$ for $n \geq 6$.*

Suppose that $OG_6 = 5\%$. Then one of the 295 LD-TTP solutions in Theorem 2 is at most 5% higher than the optimal TTP solution, found at a fraction of the computational cost. Of course, this is not necessary for the case $n = 6$ as we can use integer and constraint programming to output the TTP solution in a reasonable amount of time. However, for larger values of $n$, this linear distance relaxation technique would allow us to quickly generate close-to-optimal solutions when the exact optimal total distance is unknown or too difficult computationally. We are hopeful that this approach will help us develop better upper bounds for large unsolved benchmark instances.

## 8. Conclusion

In many professional sports leagues, teams are divided into two conferences, where teams have *intra*-league games within their own conference as well as *inter*-league games against teams from other conference. The TTP models intra-league tournament play. The NP-complete Bipartite Traveling Tournament Problem (Hoshino & Kawarabayashi, 2011) models inter-league play, and it would be interesting to see whether our linear distance relaxation can also be applied to bipartite instances to help formulate new ideas in inter-league tournament scheduling.

We conclude the paper by proposing two new benchmark instances for the Traveling Tournament Problem, as well as an open problem and a conjecture on the Linear Distance TTP. We first begin with the benchmark instances.

For each $n \geq 4$, define LINE$n$ to be the instance where the $n$ teams are located on a straight line, with a distance of one unit separating each pair of adjacent teams, i.e., $d_k = 1$ for all $1 \leq k \leq n-1$. And define INCR$n$ to be the increasing-distance scenario where the $n$ teams are arranged so that $d_k = k$ for all $1 \leq k \leq n-1$. Figure 6 illustrates the location of each team in INCR6.

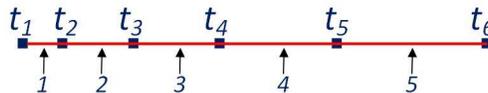

Figure 6: The instance INCR6.

By definition, the TTP solution matches the LD-TTP solution for each of these two instances. By Theorem 1, the optimal solutions for LINE6 and INCR6 have total distance 84 and 250, respectively. This naturally motivates the following problem:

**Problem 2.** *Solve the TTP for the instances LINE$n$ and INCR$n$, for $n \geq 8$.*

We conclude with one more problem, inspired by Theorem 2 which listed all seven possible optimal distances for the 6-team LD-TTP:

**Problem 3.** Let $PD_n$ denote the number of possible distances that can be a solution to the n-team LD-TTP. For example, $PD_4 = 1$ and $PD_6 = 7$. Prove or disprove that $PD_n$ is exponential in n.

## Acknowledgments

This research has been partially supported by the Japan Society for the Promotion of Science (Grant-in-Aid for Scientific Research), the C & C Foundation, the Kayamori Foundation, and the Inoue Research Award for Young Scientists. The authors thank Brett Stevens from Carleton University for suggesting the idea of the Linear Distance TTP during the 2010 Winter Meeting of the Canadian Mathematical Society.

## Appendix A

We used Maplesoft (www.maplesoft.com) to generate the set of optimal LD-TTP schedules for $n = 4$ and $n = 6$. In this appendix, we explain the process by which we generated the 36 optimal schedules for the case $n = 4$.

To simplify notation, we used the numbers 1 to 4 to represent the team numbers of opponents for road games, and the numbers 11 to 14 to represent the team numbers of opponents for home games. Thus, in our notation, the schedule on the left (from Table 2) is identical to the $4 \times 6$ matrix on the right.

| Team | 1 | 2 | 3 | 4 | 5 | 6 |
|---|---|---|---|---|---|---|
| $t_1$ | $\mathbf{t_4}$ | $\mathbf{t_3}$ | $t_2$ | $t_4$ | $t_3$ | $\mathbf{t_2}$ |
| $t_2$ | $\mathbf{t_3}$ | $\mathbf{t_4}$ | $\mathbf{t_1}$ | $t_3$ | $t_4$ | $t_1$ |
| $t_3$ | $t_2$ | $t_1$ | $\mathbf{t_4}$ | $\mathbf{t_2}$ | $\mathbf{t_1}$ | $t_4$ |
| $t_4$ | $t_1$ | $t_2$ | $t_3$ | $\mathbf{t_1}$ | $\mathbf{t_2}$ | $\mathbf{t_3}$ |

```
14 13  2  4  3 12
13 14 11  3  4  1
 2  1 14 12 11  4
 1  2  3 11 12 13
```

To produce the set of 36 schedules, the following code was used:

```
restart: with(combinat):

A1 := '<,>'(12, 1, 14, 3):  A2 := '<,>'(12, 1, 4, 13):
A3 := '<,>'(2, 11, 14, 3):  A4 := '<,>'(2, 11, 4, 13):
B1 := '<,>'(13, 14, 1, 2):  B2 := '<,>'(13, 4, 1, 12):
B3 := '<,>'(3, 14, 11, 2):  B4 := '<,>'(3, 4, 11, 12):
C1 := '<,>'(14, 13, 2, 1):  C2 := '<,>'(14, 3, 12, 1):
C3 := '<,>'(4, 13, 2, 11):  C4 := '<,>'(4, 3, 12, 11):
Z[{1, 2}] := d1: Z[{2, 3}] := d2: Z[{3, 4}] := d3:
Z[{1, 3}] := d1+d2: Z[{1, 4}] := d1+d2+d3: Z[{2, 4}] := d2+d3:

dist := proc (myinput, k)
local i, myseq, x; x := 0; myseq := [7, op(myinput), 7];
for i to 7 do
 if 7<= myseq[i] and 7<= myseq[i+1] then x:=x
 elif 7<= myseq[i] and myseq[i+1]<7 then x:=x+Z[{myseq[i+1],k}]
 elif myseq[i]<7 and 7<= myseq[i+1] then x:=x+Z[{myseq[i],k}]
```

```
  elif myseq[i]<7 and myseq[i+1]<7 then x:=x+Z[{myseq[i+1],myseq[i]}]
  else RETURN(ERROR)
  fi:
od: x: end:

checker := proc (my720)
local k, flag, x, y, goodlist, temp; goodlist := NULL;
for k to 720 do
 temp := Matrix([seq(my720[k][t], t = 1 .. 6)]); flag := 0;
 for x to 4 do for y to 5 do
  if abs(temp[x][y]-temp[x][y+1])=10 then flag:=1 fi:
 od: od:
 for x to 4 do
  if dist([seq(temp[x,k],k = 1..6)],x)<>2*(d1+d2+d3) then flag := 1 fi:
 od:
 if flag = 0 then goodlist := goodlist, temp: fi:
od: goodlist: end:

my720 := permute([A1, A4, B1, B4, C1, C4]): set1 := checker(my720):
my720 := permute([A1, A4, B1, B4, C2, C3]): set2 := checker(my720):
my720 := permute([A1, A4, B2, B3, C1, C4]): set3 := checker(my720):
my720 := permute([A1, A4, B2, B3, C2, C3]): set4 := checker(my720):
my720 := permute([A2, A3, B1, B4, C1, C4]): set5 := checker(my720):
my720 := permute([A2, A3, B1, B4, C2, C3]): set6 := checker(my720):
my720 := permute([A2, A3, B2, B3, C1, C4]): set7 := checker(my720):
my720 := permute([A2, A3, B2, B3, C2, C3]): set8 := checker(my720):
finallist := [set1, set2, set3, set4, set5, set6, set7, set8];
```

## Appendix B

We now provide the Maplesoft code from which we generated the 295 non-isomorphic schedules in Theorem 2. Due to symmetry, we only need to consider the cases $S_1, S_2, S_3, S_4$. The authors would be happy to provide the full set of 295 schedules (available as a simple .txt file upon request), and/or answer any questions that explain why this code generates the complete set of optimal schedules for the $n=6$ case of the LD-TTP.

```
restart: with(combinat):

Z := Matrix(6, 6, 0):
Z[1, 2] := a: Z[1, 3] := a+b: Z[1, 4] := a+b+c: Z[1, 5] := a+b+c+d:
Z[1, 6] := a+b+c+d+e: Z[2, 3] := b: Z[2, 4] := b+c: Z[2, 5] := b+c+d:
Z[2, 6] := b+c+d+e: Z[3, 4] := c: Z[3, 5] := c+d: Z[3, 6] := c+d+e:
Z[4, 5] := d: Z[4, 6] := d+e: Z[5, 6] := e:
for i to 6 do for j from i+1 to 6 do Z[j, i] := Z[i, j] od: od:

all252 := choose(10, 5): combos := []:
```

```
for i to 252 do
 test := all252[i]: flag := 0:
 for j to 2 do if test[j+3]-test[j] <= 3 then flag := 1: fi: od:
 for j to 4 do if test[j+1]-test[j] >= 5 then flag := 1: fi: od:
 if 'or'(test[1] >= 5, test[5] <= 6) then flag := 1 fi:
 if flag = 0 then combos := [op(combos), test]: fi:
od:

totaldist := proc (myinput, k)
local i, myseq, y; y := 0; myseq := [7, op(myinput), 7];
for i to 11 do
 if 7<= myseq[i] and 7<= myseq[i+1] then y:=y
 elif 7<= myseq[i] and myseq[i+1]<7 then y:=y+Z[myseq[i+1],k]
 elif myseq[i]<7 and 7<= myseq[i+1] then y:=y+Z[myseq[i],k]
 elif myseq[i]<7 and myseq[i+1]<7 then y:=y+Z[myseq[i+1],myseq[i]]
 else RETURN(ERROR)
 fi:
od: y: end:

getseq := proc (myfive, k)
local myperm, myseq, mylist, i, j;
mylist := []; myperm := permute('minus'({1, 2, 3, 4, 5, 6}, {k}));
for i to 120 do myseq := [seq(7, i = 1 .. 10)];
 for j to 5 do myseq[myfive[j]] := myperm[i][j]: od:
 mylist := [op(mylist), myseq]
od:
mylist: end:

checkdup := proc (tryj, j, tryk, k)
local i, val1, val2, x; x := 0;
i:=0: while x=0 and i<10 do
 i:=i+1; if tryj[i]=tryk[i] and tryj[i]<7 then x:=1: fi: od:
i:=0: while x=0 and i<10 do
 i:=i+1; if tryj[i]=k then if tryk[i]<7 then x:=1: fi: fi: od:
i:=0: while x=0 and i<10 do
 i:=i+1; if tryk[i]=j then if tryj[i]<7 then x:=1: fi: fi: od:
i:=0: while x=0 and i<10 do
 i:=i+1; if tryk[i]=j then val1:=i fi: if tryj[i]=k then val2:=i: fi: od:
if x = 0 then if abs(val1-val2) <= 1 then x := 1: fi: fi:
x: end:

fivetuple := proc (myset)
[coeff(myset,a),coeff(myset,b),coeff(myset,c),coeff(myset,d),coeff(myset,e)]:
end:
```

```
for p to 5 do for q to 5 do for r to 5 do
for s to 5 do for t to 5 do for k to 6 do
 S[k, [2*p, 2*q, 2*r, 2*s, 2*t]] := NULL:
od: od: od: od: od: od:
for kk to 6 do allvals[kk] := {}: od:

for r to 194 do x := getseq(combos[r], 1):
for s to 120 do if `in`(2, {seq(x[s][k], k = 6 .. 10)}) then
y := fivetuple(totaldist(x[s], 1));
allvals[1] := {y, op(allvals[1])}; S[1, y] := S[1, y], x[s] fi: od: od:

for r to 194 do x := getseq(combos[r], 2):
for s to 120 do y := fivetuple(totaldist(x[s], 2));
allvals[2] := {y, op(allvals[2])}; S[2, y] := S[2, y], x[s] od: od:

for r to 194 do x := getseq(combos[r], 3);
for s to 120 do y := fivetuple(totaldist(x[s], 3));
allvals[3] := {y, op(allvals[3])}; S[3, y] := S[3, y], x[s] od: od:

for r to 194 do x := getseq(combos[r], 4);
for s to 120 do y := fivetuple(totaldist(x[s], 4));
allvals[4] := {y, op(allvals[4])}; S[4, y] := S[4, y], x[s] od: od:

for r to 194 do x := getseq(combos[r], 5);
for s to 120 do y := fivetuple(totaldist(x[s], 5));
allvals[5] := {y, op(allvals[5])}; S[5, y] := S[5, y], x[s] od: od:

for r to 194 do x := getseq(combos[r], 6);
for s to 120 do y := fivetuple(totaldist(x[s], 6));
allvals[6] := {y, op(allvals[6])}; S[6, y] := S[6, y], x[s] od: od:

for pp to 10 do for qq to 10 do for rr to 10 do for ss to 10 do
 triplet1[[2*pp, 2*qq, 2*rr, 2*ss, 6]] := []: od: od: od: od:

for pp to 10 do for qq to 10 do for rr to 10 do for ss to 10 do
 triplet2[[6, 2*ss, 2*rr, 2*qq, 2*pp]] := []: od: od: od: od:

for pp to nops(allvals[1]) do
for qq to nops(allvals[2]) do
for rr to nops(allvals[3]) do
 val := allvals[1][pp]+allvals[2][qq]+allvals[3][rr];
 triplet1[val] := [op(triplet1[val]), [pp, qq, rr]]
od: od: od:

for pp to nops(allvals[4]) do
```

```
for qq to nops(allvals[5]) do
for rr to nops(allvals[6]) do
 val := allvals[4][pp]+allvals[5][qq]+allvals[6][rr];
 triplet2[val] := [op(triplet2[val]), [pp, qq, rr]]
od: od: od:

getnext := proc (inputset, x, setx)
local i, k1, k2, k3, candx, mylist;
mylist := NULL;
k1 := op('minus'({1, 2, 3, 4, 5, 6, 7}, {op(inputset[1])}));
k2 := op('minus'({1, 2, 3, 4, 5, 6, 7}, {op(inputset[2])}));
k3 := op('minus'({1, 2, 3, 4, 5, 6, 7}, {op(inputset[3])}));
for i to nops(setx) do candx := setx[i];
 if checkdup(candx, x, inputset[1], k1) = 0 then
 if checkdup(candx, x, inputset[2], k2) = 0 then
 if checkdup(candx, x, inputset[3], k3) = 0 then
 mylist := mylist, candx fi: fi: fi: od:
[mylist]: end:

getpos := proc (aval, bval, cval, dval)
local pos3, pos4, pos5, pos6;
if aval = 3 then pos3 := 2 elif aval = 4 then pos4 := 2 elif aval = 5 then
 pos5 := 2 elif aval = 6 then pos6 := 2 else RETURN(ERROR) end if;
if bval = 3 then pos3 := 3 elif bval = 4 then pos4 := 3 elif bval = 5 then
 pos5 := 3 elif bval = 6 then pos6 := 3 else RETURN(ERROR) end if;
if cval = 3 then pos3 := 7 elif cval = 4 then pos4 := 7 elif cval = 5 then
 pos5 := 7 elif cval = 6 then pos6 := 7 else RETURN(ERROR) end if;
if dval = 3 then pos3 := 8 elif dval = 4 then pos4 := 8 elif dval = 5 then
 pos5 := 8 elif dval = 6 then pos6 := 8 else RETURN(ERROR) end if;
[pos3, pos4, pos5, pos6]: end:

firsttwo := proc (aval, bval, cval, dval,new1,new2)
local pairs12, p, q, i, t1, t2, flag; pairs12 := {};
for p to nops(new1) do for q to nops(new2) do
 if checkdup(new1[p], 1, new2[q], 2) = 0 then
 if new1[p][4] <> bval and new1[p][6] <> cval and new1[p][9] <> dval and
    new2[q][2] <> aval and new2[q][5] <> bval and new2[q][7] <> cval then
 flag := 0;
 t1:=[2,aval,bval,new1[p][4],new1[p][5],new1[p][6],cval,dval,new1[p][9],2];
 t2:=[1,new2[q][2],aval,bval,new2[q][5],new2[q][6],new2[q][7],cval,dval,1];
 for i to 9 do if t1[i]=t2[i+1] and t1[i+1]=t2[i] then flag:=1 fi: od:
 if flag = 0 then pairs12 := {op(pairs12), [new1[p], new2[q]]} fi: fi: fi:
od: od:
pairs12: end:
```

```
firstthree := proc (pairs12, sixpos, new6)
local last6, mytry, trips126, p, q, k; last6 := {}; trips126 := {};
for k to nops(new6) do mytry := new6[k];
 if mytry[sixpos] = 1 and mytry[sixpos+1] = 2 then
 last6 := {op(last6), mytry} fi:
od:
for p to nops(pairs12) do for q to nops(last6) do
 if checkdup(pairs12[p][1], 1, last6[q], 6) = 0 and
  checkdup(pairs12[p][2], 2, last6[q], 6) = 0 then
 trips126 := {op(trips126), [op(pairs12[p]), last6[q]]}: fi:
od: od:
trips126: end:

steps126 := proc (new1, new2, new6)
local i, j, k, finalsol; finalsol := NULL;
for i to nops(new1) do for j to nops(new2) do
 if checkdup(new1[i], 1, new2[j], 2) = 0 then
  for k to nops(new6) do
   if checkdup(new1[i], 1, new6[k], 6) = 0 and
   checkdup(new2[j], 2, new6[k], 6) = 0 then
   finalsol := finalsol, [new1[i], new2[j], new6[k]]:
  fi: od: fi: od: od:
[finalsol]: end:

steps345 := proc (iset, new3, new4, new5)
local mylist, tryd, trye, tryf, candd, cande, candf, a, b, c;
mylist := NULL; tryd := getnext(iset, 3, new3);
if tryd <> [] then trye := getnext(iset, 4, new4);
if trye <> [] then tryf := getnext(iset, 5, new5);
if tryf <> [] then for a to nops(tryd) do candd := tryd[a];
for b to nops(trye) do cande := trye[b];
if checkdup(candd, 3, cande, 4) = 0 then
for c to nops(tryf) do candf := tryf[c];
if checkdup(candf, 5, candd, 3) = 0 then
if checkdup(candf, 5, cande, 4) = 0 then
mylist := mylist, [iset[1], iset[2], candd, cande, candf, iset[3]]:
fi: fi: od: fi: od: od: fi: fi: fi:
[mylist]: end:

allsix := proc (my126,aval,bval,cval,dval,new3,new4,new5)
local last3, last4, last5, k, mytry, tempval, pos3, pos4, pos5,
finalresult, p, q, r, s, trips345, my345, valnext, finalans;
last3:={}; last4:={}; last5:={}; finalresult:={}; trips345:={};
tempval := getpos(aval, bval, cval, dval);
pos3 := tempval[1]; pos4 := tempval[2]; pos5 := tempval[3];
```

```
for k to nops(new3) do mytry := new3[k];
 if mytry[pos3] = 1 and mytry[pos3+1] = 2 then
 if checkdup(my126[1], 1, mytry, 3) = 0 and
  checkdup(my126[2], 2, mytry, 3) = 0 and
  checkdup(my126[3], 6, mytry, 3) = 0 then
  last3 := {op(last3), mytry} fi: fi:
od:
for k to nops(new4) do mytry := new4[k];
 if mytry[pos4] = 1 and mytry[pos4+1] = 2 then
 if checkdup(my126[1], 1, mytry, 4) = 0 and
  checkdup(my126[2], 2, mytry, 4) = 0 and
  checkdup(my126[3], 6, mytry, 4) = 0 then
  last4 := {op(last4), mytry} fi: fi:
od:
for k to nops(new5) do mytry := new5[k];
 if mytry[pos5] = 1 and mytry[pos5+1] = 2 then
 if checkdup(my126[1], 1, mytry, 5) = 0 and
  checkdup(my126[2], 2, mytry, 5) = 0 and
  checkdup(my126[3], 6, mytry, 5) = 0 then
  last5 := {op(last5), mytry} fi: fi:
od:
for p to nops(last3) do for q to nops(last4) do
 if checkdup(last3[p], 3, last4[q], 4) = 0 then
  for r to nops(last5) do
   if checkdup(last3[p], 3, last5[r], 5) = 0 and
    checkdup(last4[q], 4, last5[r], 5) = 0 then
    trips345 := {op(trips345), [last3[p], last4[q], last5[r]]}:
fi: od: fi: od: od:
for s to nops(trips345) do my345 := trips345[s];
finalresult := {op(finalresult),
 [my126[1], my126[2], my345[1], my345[2], my345[3], my126[3]]}:
od:
finalresult: end:

checkallsolutions := proc(sixtuples)
local k,rr,mysolutions,cand1,cand2,cand3,cand4,cand5,cand6,
new1,new2,new3,new4,new5,new6,mytry,firsthalf,val:
mysolutions := []:
for rr to nops(sixtuples) do
 cand1 := [S[1, allvals[1][sixtuples[rr][1]]]];
 cand2 := [S[2, allvals[2][sixtuples[rr][2]]]];
 cand3 := [S[3, allvals[3][sixtuples[rr][3]]]];
 cand4 := [S[4, allvals[4][sixtuples[rr][4]]]];
 cand5 := [S[5, allvals[5][sixtuples[rr][5]]]];
 cand6 := [S[6, allvals[6][sixtuples[rr][6]]]];
```

```
 new1 := {}; new2 := {}; new3 := {}; new4 := {}; new5 := {}; new6 := {};
 for k to nops(cand1) do mytry := cand1[k]; new1 := {mytry, op(new1)}: od:
 for k to nops(cand2) do mytry := cand2[k]; new2 := {mytry, op(new2)}: od:
 for k to nops(cand3) do mytry := cand3[k]; new3 := {mytry, op(new3)}: od:
 for k to nops(cand4) do mytry := cand4[k]; new4 := {mytry, op(new4)}: od:
 for k to nops(cand5) do mytry := cand5[k]; new5 := {mytry, op(new5)}: od:
 for k to nops(cand6) do mytry := cand6[k]; new6 := {mytry, op(new6)}: od:
 firsthalf := steps126(new1, new2, new6);
 for k to nops(firsthalf) do val:=steps345(firsthalf[k],new3,new4,new5);
 if val <> [] then mysolutions := [op(mysolutions), op(val)]: fi:
od: od:
mysolutions: end:

generatesolutions := proc(sixtuples)
local cand1,cand2,cand3,cand4,cand5,cand6,new1,new2,new3,new4,new5,new6,
 k,rr,x,y,mytry,flag,aval,bval,cval,dval,pairs12,sixpos,trips126,my24,sols:
 sols := {}: my24 := permute([3,4,5,6]):
for rr to nops(sixtuples) do
 cand1 := [S[1, allvals[1][sixtuples[rr][1]]]];
 cand2 := [S[2, allvals[2][sixtuples[rr][2]]]];
 cand3 := [S[3, allvals[3][sixtuples[rr][3]]]];
 cand4 := [S[4, allvals[4][sixtuples[rr][4]]]];
 cand5 := [S[5, allvals[5][sixtuples[rr][5]]]];
 cand6 := [S[6, allvals[6][sixtuples[rr][6]]]];
 new1:={}; new2:={}; new3:={}; new4:={}; new5:={}; new6:={};
 for k to nops(cand1) do mytry := cand1[k];
  if {mytry[1], mytry[2], mytry[3], mytry[7], mytry[8]} = {7} and
  mytry[10] = 2 then new1 := {mytry, op(new1)}: fi: od:
 for k to nops(cand2) do mytry := cand2[k];
  if {mytry[3], mytry[4], mytry[8], mytry[9], mytry[10]} = {7} and
  mytry[1] = 1 then new2 := {mytry, op(new2)}: fi: od:
 for k to nops(cand3) do mytry := cand3[k]; flag := 0;
  if `and`(mytry[1] > 2, mytry[10] > 2) then
  if mytry[2] = 1 and mytry[3] = 2 then flag := 1 fi:
  if mytry[3] = 1 and mytry[4] = 2 then flag := 1 fi:
  if mytry[7] = 1 and mytry[8] = 2 then flag := 1 fi:
  if mytry[8] = 1 and mytry[9] = 2 then flag := 1 fi:
  if flag = 1 then new3 := {mytry, op(new3)}: fi: fi: od:
 for k to nops(cand4) do mytry := cand4[k]; flag := 0;
  if `and`(mytry[1] > 2, mytry[10] > 2) then
  if mytry[2] = 1 and mytry[3] = 2 then flag := 1 fi:
  if mytry[3] = 1 and mytry[4] = 2 then flag := 1 fi:
  if mytry[7] = 1 and mytry[8] = 2 then flag := 1 fi:
  if mytry[8] = 1 and mytry[9] = 2 then flag := 1 fi:
  if flag = 1 then new4 := {mytry, op(new4)}: fi: fi: od:
```

```
 for k to nops(cand5) do mytry := cand5[k]; flag := 0;
  if 'and'(mytry[1] > 2, mytry[10] > 2) then
  if mytry[2] = 1 and mytry[3] = 2 then flag := 1 fi:
  if mytry[3] = 1 and mytry[4] = 2 then flag := 1 fi:
  if mytry[7] = 1 and mytry[8] = 2 then flag := 1 fi:
  if mytry[8] = 1 and mytry[9] = 2 then flag := 1 fi:
  if flag = 1 then new5 := {mytry, op(new5)}: fi: fi: od:
 for k to nops(cand6) do mytry := cand6[k]; flag := 0;
  if 'and'(mytry[1] > 2, mytry[10] > 2) then
  if mytry[2] = 1 and mytry[3] = 2 then flag := 1 fi:
  if mytry[3] = 1 and mytry[4] = 2 then flag := 1 fi:
  if mytry[7] = 1 and mytry[8] = 2 then flag := 1 fi:
  if mytry[8] = 1 and mytry[9] = 2 then flag := 1 fi:
  if flag = 1 then new6 := {mytry, op(new6)}: fi: fi: od:
 for x to 24 do
  aval := my24[x][1]; bval := my24[x][2];
  cval := my24[x][3]; dval := my24[x][4];
  pairs12 := firsttwo(aval, bval, cval, dval,new1,new2);
  sixpos := getpos(aval, bval, cval, dval)[4];
  trips126 := firstthree(pairs12, sixpos, new6);
  for y to nops(trips126) do
   sols := {op(sols),op(allsix(trips126[y],
    aval,bval,cval,dval,new3,new4,new5))}:
  od:
 od:
od:
sols: end:

S4cases := []:
for pp to 8 do for qq to 8 do
 xx := triplet1[[8, 10, 2*pp, 2*qq, 6]];
 yy := triplet2[[6, 6, 20-2*pp, 22-2*qq, 8]];
 for u in xx do for v in yy do S4cases := [op(S4cases), [op(u), op(v)]]:
od: od: od: od:
SolutionsForS4 := generatesolutions(S4cases):

S3cases := []:
for pp from 2 to 8 do for qq to 8 do
 xx := triplet1[[8, 10, 2*pp, 2*qq, 6]];
 yy := triplet2[[6, 6, 22-2*pp, 18-2*qq, 8]];
 for u in xx do for v in yy do S3cases := [op(S3cases), [op(u), op(v)]]:
od: od: od: od:
SolutionsForS3 := generatesolutions(S3cases):
```

```
S2cases := []:
for pp to 9 do for qq to 9 do
 xx := triplet1[[10, 10, 2*pp, 2*qq, 6]];
 yy := triplet2[[6, 6, 20-2*pp, 18-2*qq, 8]];
 for u in xx do for v in yy do S2cases := [op(S2cases), [op(u), op(v)]]:
od: od: od: od:
SolutionsForS2 := checkallsolutions(S2cases):

S1cases := []:
for pp to 8 do for qq to 8 do for rr to 8 do
 xx := triplet1[[8, 2*pp, 2*qq, 2*rr, 6]]:
 yy := triplet2[[6, 18-2*pp, 20-2*qq, 18-2*rr, 8]]:
 for u in xx do for v in yy do S1cases := [op(S1cases), [op(u), op(v)]]:
od: od: od: od: od:
SolutionsForS1 := checkallsolutions(S1cases):
```